\documentclass[final,journal,10pt,twocolumn]{IEEEtran}
\usepackage{cite}
\usepackage{url}
\usepackage{lipsum}
\usepackage{siunitx}
\usepackage{pifont}
\usepackage[pdftex]{graphicx}
\usepackage{amsmath}
\usepackage{upgreek}
\usepackage{amsfonts}
\usepackage{makecell}
\usepackage{multirow}
\usepackage{booktabs}
\usepackage{tabularx}
\usepackage[usenames,dvipsnames]{xcolor} 
\usepackage{colortbl}
\usepackage{algorithm}
\usepackage{pgffor}
\usepackage{algorithmicx}
\usepackage{algpseudocode} %
\usepackage[export]{adjustbox}
\usepackage[short]{optidef}
\usepackage{soul}
\usepackage{ifthen}

\newlength\myboxwidth
\definecolor{myblue}{RGB}{230,0,0}
\graphicspath{../figures/}
\newcommand{\alphaterm}[1]{$\lambda_{#1\%}$}

\newcommand{\expGrowth}{$\lambda_{G}$} %$n_{\alpha}(t)=10^{-2}e^{\frac{t}{\tau}}$
\newcommand{\expDecay}{$\lambda_{D}$} %$n_{\alpha}(t)=e^{\frac{t}{-\tau}}$

\usepackage{tikz}
\usepackage{pgfplots}
\usepackage{pgfplotstable}
\usepackage{mathtools}
\usepackage{color,soul}
\usepackage{xcolor}
\usepgfplotslibrary{colorbrewer}
\pgfplotsset{cycle list/Set1-9}
\tikzset{every picture/.style={line width=1pt}}
\usetikzlibrary{patterns}
\usepgfplotslibrary{fillbetween}
\usepackage[eulergreek]{sansmath}
\pgfplotsset{
  tick label style = {font=\sansmath\sffamily\footnotesize},
  every axis label = {font=\sansmath\sffamily\footnotesize},
  y label style={at={(0.05,0.5)}},
  legend style = {font=\sansmath\sffamily\scriptsize},
  label style = {font=\sansmath\sffamily\footnotesize}
}
\pgfplotsset{compat=1.3} %
\DeclareMathAlphabet\mathbfcal{OMS}{cmsy}{b}{n}

\usepackage[most]{tcolorbox}
\tcbset{top=0.01mm,left=0.01mm,bottom=0.01mm,right=0.01mm}
\usepackage{efbox}
\efboxsetup{linecolor=green,linewidth=10pt}

\usepackage{booktabs,colortbl}
\pgfplotscreateplotcyclelist{sensitivity}{
teal,every mark/.append style={fill=teal!80!white},mark=*\\
orange,every mark/.append style={fill=orange!80!white},mark=square*\\
Purple!60!black,every mark/.append style={fill=Purple!80!white},mark=otimes*\\
Rhodamine!90!white,mark=star\\
lime!60!black,every mark/.append style={fill=lime!40!black},mark=diamond*\\
WildStrawberry,dashed,every mark/.append style={solid,fill=WildStrawberry!80!white},mark=*\\
brown,solid,mark=none\\
Cyan,solid,mark=none\\
Mahogany,every mark/.append style={solid,fill=gray!60!white},mark=otimes*\\
blue!80!black,dashed,mark=star,every mark/.append style=solid\\
Cyan,line width=2pt,solid,every mark/.append style={fill=black},mark=none\\
red,dashed,every mark/.append style={solid,fill=red!80!white},mark=pentagon\\
OliveGreen!60!black,,every mark/.append style={solid,fill=OliveGreen!80!white},mark=triangle\\
}

\newcolumntype{F}{>{\centering\arraybackslash}p{0.79cm}}% 0.86 a centered fixed-width-column
\newcolumntype{C}{>{\centering\arraybackslash}p{1.1cm}}% a centered fixed-width-column
\newcolumntype{P}{>{\centering\arraybackslash}p{1.1cm}}% a centered fixed-width-column
\newcolumntype{N}{>{\centering\arraybackslash}p{0.9cm}}% a centered fixed-width-column
\newcolumntype{G}{>{\centering\arraybackslash}p{1.5cm}}% a centered fixed-width-column
\newcommand{\BENmetric}{NDCG@30}
\newcommand{\DLRSDmetric}{NDCG@20}
\newcommand{\UCMmetric}{NDCG@20}
\newcommand{\BENcolname}[2]{#1-#2-\BENmetric} %F1Score
\newcommand{\DLRSDcolname}[2]{#1-#2-\DLRSDmetric}
\newcommand{\UCMcolname}[2]{#1-#2-\UCMmetric}
\pgfplotstableset{
columns/noise_prcnt/.style={column name=\textsc{SLNIR},column type={|N|}},
columns/\UCMcolname{Cls-DRL-EarlyLearningRegularization-v3}{10}/.style={column name=\textsc{ELR},precision=1,fixed zerofill},
columns/\UCMcolname{Cls-DRL-Focal-v3}{10}/.style={column name=\textsc{FL},precision=1,fixed zerofill},
columns/\UCMcolname{Cls-DRL-JoCoR-v3}{10}/.style={column name=\textsc{JoCoR},precision=1,fixed zerofill},
columns/\UCMcolname{Cls-DRL-v3}{10}/.style={column name=\textsc{CE},precision=1,fixed zerofill},
columns/\UCMcolname{NR-Cls-DRL-v3}{10}/.style={column name=\alphaterm{10},precision=1,fixed zerofill},
columns/\UCMcolname{NR-Cls-DRL-v3}{20}/.style={column name=\alphaterm{10},precision=1,fixed zerofill},
columns/\UCMcolname{NR-Cls-DRL-v3}{30}/.style={column name=\alphaterm{10},precision=1,fixed zerofill},
columns/\UCMcolname{Cls-DRL-SelfAdaptiveTraining-v3}{10}/.style={column name=\textsc{SAT},precision=1,fixed zerofill},
columns/\DLRSDcolname{Cls-DRL-EarlyLearningRegularization}{10}/.style={column name=\textsc{ELR},precision=1,fixed zerofill},
columns/\DLRSDcolname{Cls-DRL-Focal}{10}/.style={column name=\textsc{FL},precision=1,fixed zerofill},
columns/\DLRSDcolname{Cls-DRL-AsymmetricLoss}{10}/.style={column name=\textsc{ASL},precision=1,fixed zerofill},
columns/\DLRSDcolname{Cls-DRL-SelfAdaptiveTraining-v3}{10}/.style={column name=\textsc{SAT},precision=1,fixed zerofill},
columns/\DLRSDcolname{Cls-DRL-JoCoR}{10}/.style={column name=\textsc{JoCoR},precision=1,fixed zerofill},
columns/\DLRSDcolname{Cls-DRL}{10}/.style={column name=\textsc{BCE},precision=1,fixed zerofill},
columns/\DLRSDcolname{GT-DRL}{10}/.style={column name=\textsc{RRL},precision=1,fixed zerofill},
columns/\DLRSDcolname{Cls-DRL-EarlyLearningRegularization-v3}{10}/.style={column name=\textsc{ELR},precision=1,fixed zerofill},
columns/\DLRSDcolname{Cls-DRL-Focal-v3}{10}/.style={column name=\textsc{FL},precision=1,fixed zerofill},
columns/\DLRSDcolname{Cls-DRL-AsymmetricLoss-v3}{10}/.style={column name=\textsc{ASL},precision=1,fixed zerofill},
columns/\DLRSDcolname{Cls-DRL-JoCoR-v3}{10}/.style={column name=\textsc{JoCoR},precision=1,fixed zerofill},
columns/\DLRSDcolname{Cls-DRL-v3}{10}/.style={column name=\textsc{BCE},precision=1,fixed zerofill},
columns/\DLRSDcolname{GT-DRL-v3}{10}/.style={column name=\textsc{RRL},precision=1,fixed zerofill},
columns/\DLRSDcolname{NR-Cls-DRL-v2}{10}/.style={column name=\textsc{\gridcls},precision=1,fixed zerofill},
columns/\DLRSDcolname{NR-GT-DRL-v2}{10}/.style={column name=\textsc{\gridgt},precision=1,fixed zerofill},
columns/\DLRSDcolname{NR-GT-DRL-v3}{20}/.style={column name=\textsc{\gridgt},precision=1,fixed zerofill},
columns/\DLRSDcolname{NR-GT-DRL-v3}{10}/.style={column name=\textsc{\gridgt},precision=1,fixed zerofill},
columns/\DLRSDcolname{NR-Cls-DRL-v2}{20}/.style={column name=\textsc{\gridcls},precision=1,fixed zerofill},
columns/\DLRSDcolname{NR-Cls-DRL-v3}{20}/.style={column name=\textsc{\gridcls},precision=1,fixed zerofill},
columns/\DLRSDcolname{NR-GT-DRL-v2}{20}/.style={column name=\textsc{\gridgt},precision=1,fixed zerofill},
columns/\DLRSDcolname{NR-GT-DRL-v2}{30}/.style={column name=\alphaterm{30},precision=1,fixed zerofill},
columns/\DLRSDcolname{NR-GT-DRL-v2}{40}/.style={column name=\alphaterm{40},precision=1,fixed zerofill},
columns/\DLRSDcolname{NR-GT-DRL-v2}{50}/.style={column name=\alphaterm{50},precision=1,fixed zerofill},
columns/\DLRSDcolname{NR-GT-DRL-v2}{60}/.style={column name=\alphaterm{60},precision=1,fixed zerofill},
columns/\DLRSDcolname{NR-GT-DRL-v2}{70}/.style={column name=\alphaterm{70},precision=1,fixed zerofill},
columns/\DLRSDcolname{NR-GT-DRL-v2}{80}/.style={column name=\alphaterm{80},precision=1,fixed zerofill},
columns/\DLRSDcolname{NR-GT-DRL-v2}{90}/.style={column name=\alphaterm{90},precision=1,fixed zerofill},
columns/\DLRSDcolname{NR-GT-DRL-v2}{ExponentialGrowth}/.style={column name=\expGrowth,precision=1,fixed zerofill},
columns/\DLRSDcolname{NR-GT-DRL-v2}{ExponentialDecay}/.style={column name=\expDecay,precision=1,fixed zerofill},
columns/\DLRSDcolname{Seg-DRL}{10}/.style={column name=\textsc{PCE},precision=1,fixed zerofill},
columns/\DLRSDcolname{Seg-DRL-v3}{10}/.style={column name=\textsc{PCE},precision=1,fixed zerofill},
columns/\DLRSDcolname{Seg-DRL-LearningWithNoiseCorrection}{10}/.style={column name=\textsc{LNC},precision=1,fixed zerofill},
columns/\DLRSDcolname{Seg-DRL-LearningWithNoiseCorrection-v3}{10}/.style={column name=\textsc{LNC},precision=1,fixed zerofill},
columns/\DLRSDcolname{NR-Seg-DRL-v2}{10}/.style={column name=\textsc{\gridseg},precision=1,fixed zerofill},
columns/\DLRSDcolname{NR-Seg-DRL-v2}{0}/.style={column name=\alphaterm{0},precision=1,fixed zerofill},
columns/\DLRSDcolname{NR-Seg-DRL-v2}{20}/.style={column name=\alphaterm{20},precision=1,fixed zerofill},
columns/\DLRSDcolname{NR-Seg-DRL-v2}{30}/.style={column name=\alphaterm{30},precision=1,fixed zerofill},
columns/\DLRSDcolname{NR-Seg-DRL-v2}{40}/.style={column name=\alphaterm{40},precision=1,fixed zerofill},
columns/\DLRSDcolname{NR-Seg-DRL-v2}{50}/.style={column name=\alphaterm{50},precision=1,fixed zerofill},
columns/\DLRSDcolname{NR-Seg-DRL-v2}{60}/.style={column name=\alphaterm{60},precision=1,fixed zerofill},
columns/\DLRSDcolname{NR-Seg-DRL-v2}{70}/.style={column name=\alphaterm{70},precision=1,fixed zerofill},
columns/\DLRSDcolname{NR-Seg-DRL-v2}{80}/.style={column name=\alphaterm{80},precision=1,fixed zerofill},
columns/\DLRSDcolname{NR-Seg-DRL-v2}{90}/.style={column name=\alphaterm{90},precision=1,fixed zerofill},
columns/\DLRSDcolname{NR-Seg-DRL-v3}{10}/.style={column name=\textsc{\gridseg},precision=1,fixed zerofill},
columns/\DLRSDcolname{NR-Seg-DRL-v3}{20}/.style={column name=\alphaterm{20},precision=1,fixed zerofill},
columns/\DLRSDcolname{NR-Seg-DRL-v3}{30}/.style={column name=\alphaterm{30},precision=1,fixed zerofill},
columns/\DLRSDcolname{NR-Seg-DRL-v3}{40}/.style={column name=\alphaterm{40},precision=1,fixed zerofill},
columns/\DLRSDcolname{NR-Seg-DRL-v3}{50}/.style={column name=\alphaterm{50},precision=1,fixed zerofill},
columns/\DLRSDcolname{NR-Seg-DRL-v3}{60}/.style={column name=\alphaterm{60},precision=1,fixed zerofill},
columns/\DLRSDcolname{NR-Seg-DRL-v3}{70}/.style={column name=\alphaterm{70},precision=1,fixed zerofill},
columns/\DLRSDcolname{NR-Seg-DRL-v3}{80}/.style={column name=\alphaterm{80},precision=1,fixed zerofill},
columns/\DLRSDcolname{NR-Seg-DRL-v3}{90}/.style={column name=\alphaterm{90},precision=1,fixed zerofill},
columns/\DLRSDcolname{NR-Cls-DRL-v3}{10}/.style={column name=\alphaterm{10},precision=1,fixed zerofill},
columns/\DLRSDcolname{NR-Cls-DRL-v3}{20}/.style={column name=\alphaterm{20},precision=1,fixed zerofill},
columns/\DLRSDcolname{NR-Cls-DRL-v3}{30}/.style={column name=\alphaterm{30},precision=1,fixed zerofill},
columns/\DLRSDcolname{NR-Cls-DRL-v3}{40}/.style={column name=\alphaterm{40},precision=1,fixed zerofill},
columns/\DLRSDcolname{NR-Cls-DRL-v3}{50}/.style={column name=\alphaterm{50},precision=1,fixed zerofill},
columns/\DLRSDcolname{NR-Cls-DRL-v3}{60}/.style={column name=\alphaterm{60},precision=1,fixed zerofill},
columns/\DLRSDcolname{NR-Cls-DRL-v3}{70}/.style={column name=\alphaterm{70},precision=1,fixed zerofill},
columns/\DLRSDcolname{NR-Cls-DRL-v3}{80}/.style={column name=\alphaterm{80},precision=1,fixed zerofill},
columns/\DLRSDcolname{NR-Cls-DRL-v3}{90}/.style={column name=\alphaterm{90},precision=1,fixed zerofill},
columns/\DLRSDcolname{NR-Cls-DRL-v2}{0}/.style={column name=\alphaterm{0},precision=1,fixed zerofill},
columns/\DLRSDcolname{NR-Cls-DRL-v2}{30}/.style={column name=\alphaterm{30},precision=1,fixed zerofill},
columns/\DLRSDcolname{NR-Cls-DRL-v2}{40}/.style={column name=\alphaterm{40},precision=1,fixed zerofill},
columns/\DLRSDcolname{NR-Cls-DRL-v2}{50}/.style={column name=\alphaterm{50},precision=1,fixed zerofill},
columns/\DLRSDcolname{NR-Cls-DRL-v2}{60}/.style={column name=\alphaterm{60},precision=1,fixed zerofill},
columns/\DLRSDcolname{NR-Cls-DRL-v2}{70}/.style={column name=\alphaterm{70},precision=1,fixed zerofill},
columns/\DLRSDcolname{NR-Cls-DRL-v2}{80}/.style={column name=\alphaterm{80},precision=1,fixed zerofill},
columns/\DLRSDcolname{NR-Cls-DRL-v2}{90}/.style={column name=\alphaterm{90},precision=1,fixed zerofill},
columns/\DLRSDcolname{NR-Cls-DRL-v2}{ExponentialGrowth}/.style={column name=\expGrowth,precision=1,fixed zerofill},
columns/\DLRSDcolname{NR-Cls-DRL-v2}{ExponentialDecay}/.style={column name=\expDecay,precision=1,fixed zerofill},
columns/\BENcolname{Cls-DRL-EarlyLearningRegularization}{10}/.style={column name=\textsc{ELR},precision=1,fixed zerofill},
columns/\BENcolname{Cls-DRL-Focal}{10}/.style={column name=\textsc{FL},precision=1,fixed zerofill},
columns/\BENcolname{Cls-DRL-AsymmetricLoss}{10}/.style={column name=\textsc{ASL},precision=1,fixed zerofill},
columns/\BENcolname{Cls-DRL-JoCoR}{10}/.style={column name=\textsc{JoCoR},precision=1,fixed zerofill},
columns/\BENcolname{Cls-DRL}{10}/.style={column name=\textsc{BCE},precision=1,fixed zerofill},
columns/\BENcolname{GT-DRL}{10}/.style={column name=\textsc{RRL},precision=1,fixed zerofill},
columns/\BENcolname{Cls-DRL-EarlyLearningRegularization-v3}{10}/.style={column name=\textsc{ELR},precision=1,fixed zerofill},
columns/\BENcolname{Cls-DRL-Focal-v3}{10}/.style={column name=\textsc{FL},precision=1,fixed zerofill},
columns/\BENcolname{Cls-DRL-AsymmetricLoss-v3}{10}/.style={column name=\textsc{ASL},precision=1,fixed zerofill},
columns/\BENcolname{Cls-DRL-JoCoR-v3}{10}/.style={column name=\textsc{JoCoR},precision=1,fixed zerofill},
columns/\BENcolname{Cls-DRL-SelfAdaptiveTraining-v3}{10}/.style={column name=\textsc{SAT},precision=1,fixed zerofill},
columns/\BENcolname{Cls-DRL-v3}{10}/.style={column name=\textsc{BCE},precision=1,fixed zerofill},
columns/\BENcolname{GT-DRL-v3}{10}/.style={column name=\textsc{RRL},precision=1,fixed zerofill},
columns/\BENcolname{NR-Cls-DRL-v2}{10}/.style={column name=\textsc{\gridcls},precision=1,fixed zerofill},
columns/\BENcolname{NR-GT-DRL-v2}{10}/.style={column name=\textsc{\gridgt},precision=1,fixed zerofill},
columns/\BENcolname{NR-GT-DRL-v3}{10}/.style={column name=\textsc{\gridgt},precision=1,fixed zerofill},
columns/\BENcolname{NR-Cls-DRL-v2}{20}/.style={column name=\textsc{\gridcls},precision=1,fixed zerofill},
columns/\BENcolname{NR-Cls-DRL-v3}{20}/.style={column name=\textsc{\gridcls},precision=1,fixed zerofill},
columns/\BENcolname{NR-GT-DRL-v2}{20}/.style={column name=\textsc{\gridgt},precision=1,fixed zerofill},
columns/\BENcolname{NR-GT-DRL-v2}{30}/.style={column name=\alphaterm{30},precision=1,fixed zerofill},
columns/\BENcolname{NR-GT-DRL-v2}{40}/.style={column name=\alphaterm{40},precision=1,fixed zerofill},
columns/\BENcolname{NR-GT-DRL-v2}{50}/.style={column name=\alphaterm{50},precision=1,fixed zerofill},
columns/\BENcolname{NR-GT-DRL-v2}{60}/.style={column name=\alphaterm{60},precision=1,fixed zerofill},
columns/\BENcolname{NR-GT-DRL-v2}{70}/.style={column name=\alphaterm{70},precision=1,fixed zerofill},
columns/\BENcolname{NR-GT-DRL-v2}{80}/.style={column name=\alphaterm{80},precision=1,fixed zerofill},
columns/\BENcolname{NR-GT-DRL-v2}{90}/.style={column name=\alphaterm{90},precision=1,fixed zerofill},
columns/\BENcolname{NR-GT-DRL-v2}{ExponentialGrowth}/.style={column name=\expGrowth,precision=1,fixed zerofill},
columns/\BENcolname{NR-GT-DRL-v2}{ExponentialDecay}/.style={column name=\expDecay,precision=1,fixed zerofill},
columns/\BENcolname{Seg-DRL}{10}/.style={column name=\textsc{PCE},precision=1,fixed zerofill},
columns/\BENcolname{Seg-DRL-v3}{10}/.style={column name=\textsc{PCE},precision=1,fixed zerofill},
columns/\BENcolname{Seg-DRL-LearningWithNoiseCorrection}{10}/.style={column name=\textsc{LNC},precision=1,fixed zerofill},
columns/\BENcolname{Seg-DRL-LearningWithNoiseCorrection-v3}{10}/.style={column name=\textsc{LNC},precision=1,fixed zerofill},
columns/\BENcolname{NR-Seg-DRL-v2}{0}/.style={column name=\alphaterm{0},precision=1,fixed zerofill},
columns/\BENcolname{NR-Seg-DRL-v2}{10}/.style={column name=\alphaterm{10},precision=1,fixed zerofill},
columns/\BENcolname{NR-Seg-DRL-v2}{20}/.style={column name=\alphaterm{20},precision=1,fixed zerofill},
columns/\BENcolname{NR-Seg-DRL-v2}{30}/.style={column name=\alphaterm{30},precision=1,fixed zerofill},
columns/\BENcolname{NR-Seg-DRL-v2}{40}/.style={column name=\alphaterm{40},precision=1,fixed zerofill},
columns/\BENcolname{NR-Seg-DRL-v2}{50}/.style={column name=\alphaterm{50},precision=1,fixed zerofill},
columns/\BENcolname{NR-Seg-DRL-v2}{60}/.style={column name=\alphaterm{60},precision=1,fixed zerofill},
columns/\BENcolname{NR-Seg-DRL-v2}{70}/.style={column name=\alphaterm{70},precision=1,fixed zerofill},
columns/\BENcolname{NR-Seg-DRL-v2}{80}/.style={column name=\alphaterm{80},precision=1,fixed zerofill},
columns/\BENcolname{NR-Seg-DRL-v2}{90}/.style={column name=\alphaterm{90},precision=1,fixed zerofill},
columns/\BENcolname{NR-Seg-DRL-v3}{10}/.style={column name=\alphaterm{10},precision=1,fixed zerofill},
columns/\BENcolname{NR-Seg-DRL-v3}{20}/.style={column name=\alphaterm{20},precision=1,fixed zerofill},
columns/\BENcolname{NR-Seg-DRL-v3}{30}/.style={column name=\alphaterm{30},precision=1,fixed zerofill},
columns/\BENcolname{NR-Seg-DRL-v3}{40}/.style={column name=\alphaterm{40},precision=1,fixed zerofill},
columns/\BENcolname{NR-Seg-DRL-v3}{50}/.style={column name=\alphaterm{50},precision=1,fixed zerofill},
columns/\BENcolname{NR-Seg-DRL-v3}{60}/.style={column name=\alphaterm{60},precision=1,fixed zerofill},
columns/\BENcolname{NR-Seg-DRL-v3}{70}/.style={column name=\alphaterm{70},precision=1,fixed zerofill},
columns/\BENcolname{NR-Seg-DRL-v3}{80}/.style={column name=\alphaterm{80},precision=1,fixed zerofill},
columns/\BENcolname{NR-Seg-DRL-v3}{90}/.style={column name=\alphaterm{90},precision=1,fixed zerofill},
columns/\BENcolname{NR-Cls-DRL-v2}{0}/.style={column name=\alphaterm{0},precision=1,fixed zerofill},
columns/\BENcolname{NR-Cls-DRL-v2}{30}/.style={column name=\alphaterm{30},precision=1,fixed zerofill},
columns/\BENcolname{NR-Cls-DRL-v2}{40}/.style={column name=\alphaterm{40},precision=1,fixed zerofill},
columns/\BENcolname{NR-Cls-DRL-v2}{50}/.style={column name=\alphaterm{50},precision=1,fixed zerofill},
columns/\BENcolname{NR-Cls-DRL-v2}{60}/.style={column name=\alphaterm{60},precision=1,fixed zerofill},
columns/\BENcolname{NR-Cls-DRL-v2}{70}/.style={column name=\alphaterm{70},precision=1,fixed zerofill},
columns/\BENcolname{NR-Cls-DRL-v2}{80}/.style={column name=\alphaterm{80},precision=1,fixed zerofill},
columns/\BENcolname{NR-Cls-DRL-v2}{90}/.style={column name=\alphaterm{90},precision=1,fixed zerofill},
columns/\BENcolname{NR-Cls-DRL-v2}{ExponentialGrowth}/.style={column name=\expGrowth,precision=1,fixed zerofill},
columns/\BENcolname{NR-Cls-DRL-v2}{ExponentialDecay}/.style={column name=\expDecay,precision=1,fixed zerofill},
columns/\BENcolname{NR-Cls-DRL-v3}{10}/.style={column name=\alphaterm{10},precision=1,fixed zerofill},
columns/\BENcolname{NR-Cls-DRL-v3}{20}/.style={column name=\alphaterm{20},precision=1,fixed zerofill},
columns/\BENcolname{NR-Cls-DRL-v3}{30}/.style={column name=\alphaterm{30},precision=1,fixed zerofill},
columns/\BENcolname{NR-Cls-DRL-v3}{40}/.style={column name=\alphaterm{40},precision=1,fixed zerofill},
columns/\BENcolname{NR-Cls-DRL-v3}{50}/.style={column name=\alphaterm{50},precision=1,fixed zerofill},
columns/\BENcolname{NR-Cls-DRL-v3}{60}/.style={column name=\alphaterm{60},precision=1,fixed zerofill},
columns/\BENcolname{NR-Cls-DRL-v3}{70}/.style={column name=\alphaterm{70},precision=1,fixed zerofill},
columns/\BENcolname{NR-Cls-DRL-v3}{80}/.style={column name=\alphaterm{80},precision=1,fixed zerofill},
columns/\BENcolname{NR-Cls-DRL-v3}{90}/.style={column name=\alphaterm{90},precision=1,fixed zerofill},
empty cells with={--}, % replace empty cells with '--'
every head row/.style={before row=\hline\rowcolor[gray]{0.95},after row=\hline\hline},
every last row/.style={after row=\hline},
every first column/.style={
column type/.add={|}{}
},
every last column/.style={
column type/.add={}{|}
},
every odd row/.style={
before row={\rowcolor[gray]{0.94}}},
}
\usepackage{pdftexcmds}
\newcommand{\nbDetPlotVThree}[4]{%  <--  HERE
    \begin{tikzpicture}[scale = 0.55]%
	\begin{axis}[
    height=7cm,
    width=11cm,
    grid=both,
    grid style={line width=.1pt, draw=gray!10},
    major grid style={line width=.2pt,draw=gray!50},
    minor x tick num=4,
    minor y tick num=4,
    xlabel= {\normalsize Epoch},   
    ylabel= {\normalsize \if#41$F_1$ Score\else Accuracy\fi~(\%)},
    label shift = {-5pt},
    xmin=0,xmax=100,
    cycle list name=sensitivity] %sensitivity
    \pgfplotsset{cycle list shift={\numexpr #3/10-1}}
    \addplot+[name path=capacity] table [x index={0}, y expr=\thisrowno{\if#418\else5\fi}, col sep=comma]{\datapath{noise_det/conf_matrix_noisy_selection_#1-#2-DRL-v3-DenseNet121-embed128-noise_prcnt#3-bs128-epoch100-noisy_sel_prcnt#3.csv}};
    \addplot+[line width=2pt,brown,solid,every mark/.append style={fill=black},mark=none] table [x index={0}, y expr=\thisrowno{\if#418\else5\fi}, col sep=comma]{\datapath{noise_det/conf_matrix_noisy_selection_#1-\ifnum\pdfstrcmp{#2}{NR-Seg}=0Seg\else Cls\fi-DRL-v3-TopK-DenseNet121-embed128-noise_prcnt#3-bs128-epoch100-noisy_sel_prcntTopK.csv}};
    \addplot+[line width=2pt,Cyan,solid,every mark/.append style={fill=black},mark=none] table [x index={0}, y expr=\thisrowno{\if#418\else5\fi}, col sep=comma]{\datapath{noise_det/conf_matrix_noisy_selection_#1-random-noise_prcnt#3-noisy_sel_prcnt#3.csv}};
    \end{axis}
    \end{tikzpicture}%  <--  HERE
    }
    
\newcommand{\lossname}[2]{\ifnum#1=0dis_loss_noise_prcnt#2\else\ifnum#1=1gen_loss_noise_prcnt#2\else dis_gen_loss_noise_prcnt#2\fi\fi}
\newcommand{\losscolorname}[1]{\ifnum#1=0OliveGreen!60!black\else\ifnum#1=1blue!70!teal\else red!80!white\fi\fi}
\newcommand{\losstitle}[1]{\ifnum#1=0Discriminative Task Head on $\mathcal{B}$\else\ifnum#1=1Generative Task Head on $\mathcal{B}$\else Discriminative \& Generative Task Heads on $\mathcal{C}$ \& $\mathcal{W}$ \fi\fi}%Hybrid Representation Learning
\newcommand{\losslabel}[1]{\ifnum\pdfstrcmp{#1}{Cls-DRL}=0$\mathcal{L}_{BCE}$\else$\mathcal{L}_{RRL}$ \normalsize($\times10^{-2}$)\fi}
\newcommand{\lossmul}[1]{\ifnum\pdfstrcmp{#1}{Cls-DRL}=0 1\else 100\fi}

\newcommand{\abFigureVThree}[3]{%  <--  HERE
    \begin{tikzpicture}[scale = 0.55]%0.45
	\begin{axis}[
    height=7cm,
    width=11cm,
    grid=both,
    grid style={line width=.1pt, draw=gray!10},
    major grid style={line width=.2pt,draw=gray!50},
    minor x tick num=4,
    minor y tick num=4,
    xlabel= {\normalsize SLNIR (\%)},
    ylabel= {\normalsize NDCG~(\%)},
    xmin=0,xmax=60]
    \addplot+[name path=capacity,mark=diamond,color=OliveGreen!60!black,mark options={fill=white,scale=1.5},line width=1pt] table [x=noise_prcnt, y=\csname#1metric\endcsname, col sep=comma] {\datapath{#1_#2-v3_10_val.csv}};
    \addplot+[name path=capacity,mark=x,color=blue!70!teal,mark options={fill=white,scale=1.5},line width=1pt] table [x=noise_prcnt, y=\csname#1metric\endcsname, col sep=comma] {\datapath{#1_#2-VAE-v3_10_val.csv}};
    \addplot+[name path=capacity,color=Melon,mark=triangle,mark options={fill=white,scale=1.5},line width=1pt] table [x=noise_prcnt, y=\csname#1metric\endcsname, col sep=comma] {\datapath{#1_NR-#2-v3-woSelection_10_val.csv}};
    \addplot+[name path=capacity,color=red!80!white,mark=*,mark options={fill=white,scale=1.5},line width=1pt] table [x=noise_prcnt, y=\csname#1metric\endcsname, col sep=comma] {\datapath{#1_NR-#2-v3_#3_val.csv}};
    \end{axis}
    \end{tikzpicture}%  <--  HERE
    }
\def\compilefigs{0}
\newcommand{\inputfig}[1]{\if\compilefigs1\input{sources/plots/#1}\else\includegraphics{figures/#1}\fi}
\newcommand{\inputtable}[1]{\if\compilefigs1\input{sources/plots/#1}\else\includegraphics{figures/#1}\fi}
\newcommand{\datapath}[1]{data/#1}%{../../data/#1}

\newcommand{\grid}{GRID}

\newcommand{\gridcls}{\grid~(BCE)}

\newcommand{\gridgt}{\grid~(RRL)}
\newcommand{\gridseg}{\grid~(PCE)}

\begin{document}
\title{Generative Reasoning Integrated Label Noise Robust Deep Image Representation Learning}

\author{%
    Gencer~Sumbul,~\IEEEmembership{Graduate Student Member,~IEEE},
	and~Beg{\"u}m~Demir,~\IEEEmembership{Senior Member,~IEEE}%
    \thanks{Gencer Sumbul and Beg{\"u}m Demir are with the Faculty of Electrical Engineering and Computer Science, Technische Universit\"at Berlin, 10623 Berlin, Germany, also with the BIFOLD - Berlin Institute for the Foundations of Learning and Data, 10623 Berlin, Germany.
    Email: \mbox{gencer.suembuel@tu-berlin.de}, \mbox{demir@tu-berlin.de}.}%
    % \thanks{This article has supplementary downloadable material available at \protect\url{https://doi.org/10.1109/TIP.2023.3293776}, provided by the authors.}%
}
\maketitle

\begin{abstract}
The development of deep learning based image representation learning (IRL) methods has attracted great attention for various image understanding problems. Most of these methods require the availability of a set of high quantity and quality of annotated training images, which can be time-consuming, complex and costly to gather. To reduce labeling costs, crowdsourced data, automatic labeling procedures or citizen science projects can be considered. However, such approaches increase the risk of including label noise in training data. It may result in overfitting on noisy labels when discriminative reasoning is employed as in most of the existing methods. This leads to sub-optimal learning procedures, and thus inaccurate characterization of images. To address this issue, in this paper, we introduce a generative reasoning integrated label noise robust deep representation learning (GRID) approach. The proposed GRID approach aims to model the complementary characteristics of discriminative and generative reasoning for IRL under noisy labels. To this end, we first integrate generative reasoning into discriminative reasoning through a supervised variational autoencoder. This allows the proposed GRID approach to automatically detect training samples with noisy labels. Then, through our label noise robust hybrid representation learning strategy, GRID adjusts the whole learning procedure for IRL of these samples through generative reasoning and that of the other samples through discriminative reasoning. Our approach learns discriminative image representations while preventing interference of noisy labels during training independently from the IRL method being selected. Thus, unlike the existing label noise robust methods, GRID does not depend on the type of annotation, label noise, neural network architecture, loss function or learning task, and thus can be directly utilized for various image understanding problems. Experimental results show the effectiveness of the proposed GRID approach compared to the state-of-the-art methods. The code of the proposed approach is publicly available at \url{https://github.com/gencersumbul/GRID}.
\end{abstract}
%https://git.tu-berlin.de/rsim/GRID
\begin{IEEEkeywords}
Generative reasoning, discriminative reasoning, noise robust learning, image representation learning
\end{IEEEkeywords}

\section{Introduction}
Deep learning (DL) based image representation learning (IRL) has attracted significant attention for image understanding applications. Due to its undoubted capabilities to model higher level image semantics, it is integrated into various learning tasks (e.g., multi-label classification~{\cite{Chen:2022, Gao:2021, Sumbul:2020}}, semantic segmentation~{\cite{Ding:2022, Gao:2022, Tao:2022}}, image captioning~{\cite{Cheng:2022, Sumbul:2021:3}}, change detection~{\cite{Lin:2023, Zhang:2021}} and content-based image retrieval~{\cite{Peng:2023, Sumbul:2022, Sumbul:2021}}, etc.) in many fields, e.g., natural image analysis{~\cite{Wei:2022}}, remote sensing{~\cite{Zhang:2022}}, medical imaging{~\cite{Zhou:2021}}, etc.

DL-based IRL is generally achieved in a supervised way during the optimization of a loss function based on the characteristics of a learning task. To effectively learn DL model parameters, the availability of a high quantity and quality of annotated training images is required. Depending on the considered learning task, annotations of training images can be given at scene-level or pixel-level. For scene-level annotations, each training image is annotated by either a single label, which is associated to the most significant content of the image, or multi-labels. In general, the manual collection of image annotations by domain experts for large scale data can be time consuming, complex and costly. To address this issue, crowdsourced data, automatic labeling procedures or citizen science projects can be considered. These strategies provide image annotations at zero cost. However, there can be annotation errors or the considered data source can be outdated with respect to images due to possible changes. Thus, these strategies increase the risk of including noisy labels in training data. It is worth noting that for a scene-level single-label and a pixel-level noisy annotation, label noise occurs as an incorrect label associated to an image and a pixel, respectively. However, for a scene-level multi-label noisy annotation, it can emerge as a missing label (i.e., a class is present in an image while the corresponding label is not assigned to that image), a wrong label (i.e., a class is not present in an image while its label is assigned to the image) or combination of both missing and wrong labels.

Most of the existing DL-based IRL methods employ discriminative learning (i.e., discriminative reasoning) of image representations. This is based on directly modeling a posterior data distribution $p(y|x)$ by utilizing $(x,y)$ image annotation pairs from training data. The effectiveness of the discriminative reasoning has been proven compared to generative reasoning (which is based on modelling the joint data distribution $p(x,y)$) when training data is abundant~\cite{Raina:2003}. However, discriminative models are more sensitive to label noise compared to generative models. Accordingly, discriminative learning of image representations with noisy labels may result in overfitting of the considered deep neural network (DNN) to noisy labels and lack of its generalization capability, and thus inaccurate characterization of images during both training and inference~\cite{Zhang:2017, Song:2022}.

To address this problem, several methods are presented to improve the robustness of discriminative IRL when training data includes noisy labels. All these methods are potentially effective for DL-based IRL under noisy labels. However, most of them are dependent on the type of: i) label noise present in training data; ii) image annotation; iii) loss function (e.g., cross-entropy, focal loss etc.); iv) deep neural network architecture; or v) learning task. Some methods also require the availability of a subset of the training set, which includes clean labels, or require the computationally demanding noise correction strategies prior to training. Thus, they may not be directly integrated into different IRL scenarios. 

To overcome this issue, in this paper, we introduce a Generative Reasoning Integrated Label Noise Robust Deep Representation Learning (denoted as \textbf{GRID} hereafter) approach. The proposed GRID approach aims to model the complementary characteristics of discriminative and generative reasoning for IRL under noisy labels. To this end, for discriminative reasoning, we first employ a DNN composed of an image encoder (i.e., CNN backbone) and a discriminative task head for modelling the posterior distribution of labeled images as in the most of supervised DL-based IRL methods. Then, we integrate generative reasoning into discriminative reasoning through a supervised variational autoencoder (which includes a variational encoder, a feature decoder and a generative task head) followed by the CNN backbone for modelling the joint distribution of labeled images. This allows the proposed GRID approach to automatically detect training samples with noisy labels based on the loss values acquired from discriminative and generative task heads. Then, through our label noise robust hybrid representation learning strategy, the model parameters of the considered DNN is updated through: i) generative reasoning for the samples with noisy labels; and ii) discriminative reasoning for the remaining samples in training data. Accordingly, our approach allows to learn discriminative image representations through the CNN backbone, while preventing the overfitting on noisy labels during training independent from the IRL method being selected. Thus, unlike the existing label noise robust methods, GRID does not depend on the type of annotation, label noise, DNN architecture, loss function or learning task. It also does not require a reliable subset of a training set or require a computationally demanding noise correction strategy prior to training. Thus, our approach can be directly utilized for various scenarios in IRL. In this paper, we consider two IRL scenarios, where training images are annotated with: 1) scene-level noisy multi-labels; and 2) pixel-level noisy labels. Under these scenarios, we consider three learning tasks with the corresponding loss functions and DNN architectures. For different scenarios and learning tasks, we conduct experiments on a single application for the sake of simplicity. This application is selected as content-based image retrieval (CBIR), for which accurately learning image features is crucial. We would like to note that, according to our knowledge, GRID is the first approach that combines generative and discriminative reasoning for supervised IRL under noisy labels independently of the considered IRL method that leads to accurately learning image representations while preventing interference of noisy labels during training.

The rest of the paper is organized as follows: Section II presents the related works on DL-based label noise robust IRL. Section III introduces the proposed GRID approach. Section IV describes the considered datasets and the experimental setup, while Section V provides the experimental results. Section VI concludes our paper.
\begin{figure*}[t]
    \centering
    \input{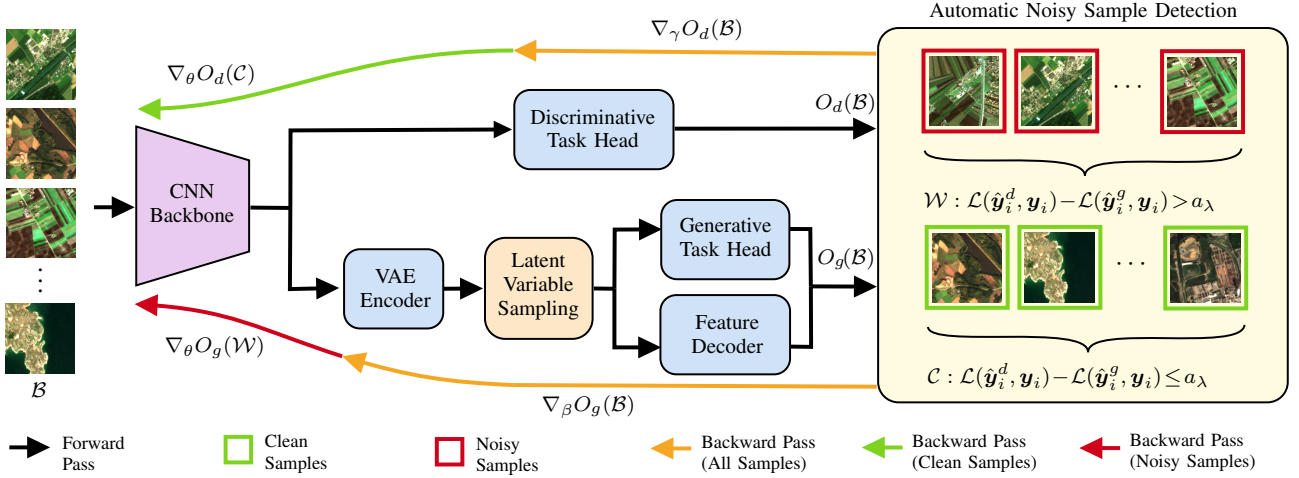}
    \caption{An illustration of the training of our GRID approach that jointly leverages the robustness of generative reasoning towards noisy labels and the effectiveness of discriminative reasoning on image representation learning. During the forward pass on a mini-batch $\mathcal{B}$, the loss values $O_d(\mathcal{B})$, $O_g(\mathcal{B})$ and the predicted labels $\hat{\mathcal{Y}}^d$, $\hat{\mathcal{Y}}^g$ are obtained through discriminative and generative reasoning for a given learning task. While discriminative reasoning is achieved based on a CNN backbone and a discriminative task head, generative reasoning is achieved based on a VAE encoder and a VAE decoder (which is composed of a feature decoder and a generative task head). Then, the set $\mathcal{W}$ of training samples with noisy labels (i.e., noisy samples) and the set $\mathcal{C}$ of training samples with correct labels (i.e., clean samples) are constructed through our automatic noisy sample detection procedure based on the values of the loss function $\mathcal{L}$ associated with the learning task. During the backward pass, the model parameters except the CNN backbone parameters are updated with all samples based on $\nabla _{\gamma } O_{d}( \mathcal{B})$ and $\nabla _{\beta } O_{g}( \mathcal{B})$. The parameters of the CNN backbone are updated through: i) the generative task head for the noisy samples based on $\nabla _{\theta } O_{g}( \mathcal{W})$; and ii) the discriminative task head for the clean samples based on $\nabla _{\theta } O_{d}( \mathcal{C})$.}
    \label{fig:approach_figure}
\end{figure*}

\section{Deep Representation Learning Methods for Images with Noisy Labels}

The recent studies for label noise robust IRL are mainly concentrated on the development of: i) deep architectures~{\cite{Yao:2019, zhang_remote_2020}}; ii) loss functions~{\cite{Lin:2020, kang_robust_2021, li_improved_2022}}; iii) regularization strategies{~\cite{Liu:2020}}; and iv) sample selection and label adjustment techniques{~\cite{Wei:2020}} mostly for single-label image scene classification problems. The methods in the first category focus on designing DNN architectures specific to training data with noisy labels. As an example, in{~\cite{Yao:2019}}, a contrastive-additive noise network is proposed to model trustworthiness of noisy labels in the context of image classification. To this end, it includes a probabilistic latent variable model as a contrastive layer to estimate the quality of labels and an additive layer to aggregate the class predictions and noisy labels. A noisy label distillation method is introduced in{~\cite{zhang_remote_2020}} to leverage the knowledge learned through a teacher model on images with noisy labels for a student model. In this method, two convolutional neural networks (CNNs) are employed as a teacher-student framework, while a clean and reliable subset of a training set is assumed to be available for the student CNN. The methods in the second category are devoted to develop loss functions, which have robust characteristics to noisy labels. For instance, the symmetric cross-entropy loss function is proposed in{~\cite{Wang:2019}} to prevent the overfitting of noisy labels for easy samples and under-learning for hard samples by combining the standard cross-entropy loss function with its reverse, where class probabilities are swapped with ground truth labels. In{~\cite{kang_robust_2021}}, a down-weighting factor is integrated into the normalized softmax loss function to reduce the effect of wrongly classified images (which are assumed to be associated with noisy labels) on the model parameter updates. The methods in the third category aim at regularizing the whole learning procedure to prevent overfitting on noisy labels. As an example, in{~\cite{Liu:2020}}, a regularization term is integrated into cross-entropy loss to guide the learning process with the class predictions from an early stage of training to prevent memorization of noisy-labels. The methods in the last category focus on first detecting images associated with correct labels or adjusting noisy labels, and then learning through those samples or adjusted labels. For instance, in{~\cite{Wei:2020}}, a joint training with co-regularization approach employs collaborative learning of two CNNs for the selection of correct labels by an agreement strategy. 

We would like to note that DL-based label noise robust IRL has been extensively studied for images annotated with single-labels in the context of scene classification problems. For a detailed review of such studies, we refer the reader to{~\cite{Song:2022}}. However, relatively less attention has been devoted to label noise robust IRL for images annotated with scene-level multi-labels and pixel-based labels. As an example for multi-label images, in{~\cite{aksoy_consensual_2022}} a collaborative learning framework is proposed in the context of multi-label classification to identify and exclude images with noisy labels during training. To this end, it employs two CNNs operating collaboratively, which are forced to characterize distinct image representations and to produce similar predictions. In{~\cite{Ridnik:2021}}, the asymmetric loss function is proposed to dynamically decrease the weights of negative classes in multi-labels that decreases the effect of images with missing labels on IRL. In{~\cite{Burgert:2022}}, the effects of different label noise types in multi-label image classification problems are investigated, while different noise robust methods are integrated from single-label to multi-label classification problems. Apart from scene-level image classification, IRL under pixel-based noisy labels has been recently studied for semantic segmentation problems~{\cite{Yang:2023, Dong:2022, Rahaman:2022, ahmed_dense_2021}}. As an example, in{~\cite{Dong:2022}} an online noise correction approach is proposed to detect and correct pixel-level noisy labels via information entropy at the early stage of training, and thus to continue training with corrected labels. In{~\cite{Yang:2023}}, a noise robust method is introduced for instance segmentation problems. In this method, potentially noisy foreground instances are defined based on the cross-entropy loss values. Then, IRL for those regions is achieved through the self-supervised contrastive and symmetric cross-entropy loss functions to prevent incorrect gradient guidance of noisy labels.

\section{Proposed Generative Reasoning Integrated Label Noise Robust Deep Representation Learning Approach}
Let $\mathcal{X}{=}\{\boldsymbol{x}_1,\ldots,\boldsymbol{x}_M\}$ be an image archive that includes $M$ images, where $\boldsymbol{x}_t$ is the $t$th image in the archive. We assume that a training set $\mathcal{T} = \{(\boldsymbol{x}_1, \boldsymbol{y}_1),\ldots,(\boldsymbol{x}_K, \boldsymbol{y}_K)\}$ that includes $K$ i.i.d samples of random variables $\mathbf{x}$ and $\mathbf{y}$ is available. $\boldsymbol{x}_i$ is the $i$th image and $\boldsymbol{y}_i$ is the corresponding image annotation. Annotations of training images can be given at pixel-level or scene-level. An image can be annotated by a broad category label (i.e., single-label) or multi-labels. We assume that the labels in the set $\mathcal{Y}$ of training image annotations can be noisy. For a scene-level single-label or a pixel-level noisy annotation, label noise may occur as an incorrect label associated to an image or a pixel, respectively. For a scene-level multi-label noisy annotation, label noise may occur as a missing label, a wrong label or combination of both missing and wrong labels.

The proposed GRID approach aims to jointly leverage the robustness of generative reasoning towards noisy labels and the effectiveness of discriminative reasoning on IRL. This is achieved by first integrating generative reasoning into discriminative reasoning through a supervised variational autoencoder, and then characterizing discriminative image representations while preventing interference of noisy labels through our label noise robust hybrid representation learning strategy. Fig. \ref{fig:approach_figure} shows an illustration of the proposed GRID approach. We first provide general information on discriminative reasoning, and then present our approach in detail in the following subsections.

\subsection{Basics on Discriminative Reasoning}
DL-based IRL methods through discriminative reasoning aim to employ the discriminative capabilities of DNNs for the characterization of image features. This is achieved by maximizing the posterior distribution of labeled images $p(\mathbf{y}|\mathbf{x})$ during training. To this end, the considered DNN typically includes an image encoder (i.e., a CNN backbone) and a discriminative task head including fully connected or convolutional layers (which is branched out from the CNN backbone). Let $\phi : \theta, \mathcal{X} \mapsto \mathcal{F}$ be any type of CNN backbone that maps the image $\boldsymbol{x}_i$ into the corresponding image descriptor $\boldsymbol{f}_i$, which is a sample of random variable $\mathbf{f}$. $\theta$ is the set of CNN parameters and $\mathcal{F}$ is the set of all descriptors for $\mathcal{X}$. Let $t^d \!:\! \gamma, \mathcal{F} \mapsto \hat{\mathcal{Y}}^d$ be a discriminative task head that maps the image descriptor into the corresponding label prediction associated with the image $\boldsymbol{x}_i$ [i.e., $t^d(\phi(\boldsymbol{x}_i;\theta);\gamma) = \hat{\boldsymbol{y}}_i^d$], where $\gamma$ is the task head parameters and $\hat{\mathcal{Y}}^d$ is the set of all predicted image labels. The CNN backbone models global image representation space, while overall DNN models the posterior distribution $p(\mathbf{y}|\mathbf{x})$. In this formulation, the model parameters $\theta \cup \gamma$ are updated to maximize $\mathbb{E}_{T}[\text{log}p(\mathbf{y}|\mathbf{x})]$. Accordingly, the objective function $O_d$ associated with discriminative reasoning for a set of samples $\mathcal{S}$ is written as follows:
\begin{equation}
    O_d(\mathcal{S}) = \frac{1}{|\mathcal{S}|}\sum_{(\boldsymbol{x}_i,\boldsymbol{y}_i)\in\mathcal{S}}\mathcal{L}(t^d(\phi(\boldsymbol{x}_i;\theta);\gamma),\boldsymbol{y}_i);\;\exists \mathcal{L} \in \mathfrak{L},
\end{equation}
where $\mathfrak{L}$ is the set of all loss functions, whose each element is capable of measuring how different the prediction $\hat{\boldsymbol{y}}_i^d$ is from $\boldsymbol{y}_i$. Accordingly, any loss function that can measure the sample-wise error can be used for $\mathcal{L}$.

The discriminative learning of image representations has been found successful for many applications when the labeled training data is abundant~\cite{Raina:2003}. However, learning image representations via modeling the posterior distribution of training data can be sensitive to noisy labels included in the calculation of the loss function. When the ratio of noisy labels over $\mathcal{Y}$ is significantly high, the considered DNN can suffer from the overfitting on noisy labels leading to inaccurate IRL and lack of the generalization capability of the considered DNN~\cite{Zhang:2017, Song:2022}.

\subsection{Integration of Generative Reasoning}
Generative learning of image representations via modeling the joint data distribution $p(\mathbf{x},\mathbf{y})$ limits the overfitting of the considered DNN on noisy labels during training. Thus, it is proven to be more robust to noisy labels compared to discriminative learning~\cite{Lee:2019}. However, learning image representations via generative reasoning may limit to accurately characterize discriminative image descriptors, and thus may lead to inaccurate IRL. Accordingly, the proposed GRID approach aims at effectively integrating generative reasoning into discriminative reasoning to achieve discriminative and generative modelling of images in a single learning procedure.

To model $p(\mathbf{x},\mathbf{y})$, we assume that $\mathbf{x}$ and $\mathbf{y}$ are generated through a latent variable $\mathbf{z}$. Each sample of the latent variable $\boldsymbol{z}_i$ is generated from a prior distribution $p(\mathbf{z})$, while $\boldsymbol{x}_i$ and $\boldsymbol{y}_i$ are generated from $p(\mathbf{x},\mathbf{y}|\mathbf{z})$. It is worth noting that the marginal likelihood over the latent variable $\int p(\mathbf{z})p(\mathbf{x},\mathbf{y}|\mathbf{z})d\mathbf{z}$ is intractable for DNNs since it is hard to find an analytical solution for the posterior distribution of the latent variable $p(\mathbf{z}|\mathbf{x},\mathbf{y})$. To this end, we utilize a variational auto-encoder (VAE) introduced in~\cite{Kingma:2014} as a latent variable model. Accordingly, we approximate the true posterior distribution of latent variable with a variational approximate posterior $q(\mathbf{z}|\mathbf{x},\mathbf{y})$ of known functional form (e.g., a Gaussian distribution parameterized by the encoder of a VAE). Then, the variational lower bound on the marginal log-likelihood (i.e., evidence lower bound [ELBO]) is defined as follows:
\begin{equation}
\label{eq:elbo1}
\begin{aligned}
\text{log}\;p_{\beta^d}(\mathbf{x},\mathbf{y}) \geq & \;\mathbb{E}_{q_{\beta^e}(\mathbf{z}|\mathbf{x},\mathbf{y})}[\text{log} \; p_{\beta^d}(\mathbf{x},\mathbf{y}|\mathbf{z})]\\
&- D_{\text{KL}}(q_{\beta^e}(\mathbf{z}|\mathbf{x},\mathbf{y})\;||\;p_{\beta^d}(\mathbf{z})),
\end{aligned}
\end{equation}
where $D_{\text{KL}}(\cdot||\cdot)$ is the Kullback-Leibler (KL) divergence~\cite{Kullback:1951}, $\beta^e$ is the VAE encoder parameters and $\beta^d$ is the VAE decoder parameters. It is worth noting that $q_{\beta^e}(\mathbf{z}|\mathbf{x})$ is a sufficient statistic for $q_{\beta^e}(\mathbf{z}|\mathbf{x},\mathbf{y})$. It guarantees that $\mathbf{z}$ generated from $\mathbf{x}$ embodies the same information when it is jointly generated from $\mathbf{x}$ and $\mathbf{y}$~\cite{Feng:2021}. Since $p_{\beta_d}(\mathbf{x},\mathbf{y}|\mathbf{z})$ can be factorized into $p_{\beta_d}(\mathbf{x}|\mathbf{z})p_{\beta_d}(\mathbf{y}|\mathbf{z})$ (i.e, conditional independence), (\ref{eq:elbo1}) can be written as follows:
\begin{equation}
\label{eq:elbo2}
\begin{aligned}
\text{log}\;p_{\beta^d}(\mathbf{x},\mathbf{y}) \geq & \;\mathbb{E}_{q_{\beta^e}(\mathbf{z}|\mathbf{x})}[\text{log} \; p_{\beta^d}(\mathbf{x}|\mathbf{z})]\\
&+\mathbb{E}_{q_{\beta^e}(\mathbf{z}|\mathbf{x})}[\text{log} \; p_{\beta^d}(\mathbf{y}|\mathbf{z})]
\\
&- D_{\text{KL}}(q_{\beta^e}(\mathbf{z}|\mathbf{x})\;||\;p_{\beta^d}(\mathbf{z})).
\end{aligned}
\end{equation}
We define the variational approximate posterior and the latent prior as multivariate Gaussian distributions as follows:
\begin{equation}
    \boldsymbol{z}_i \sim q_{\beta^e}(\mathbf{z}|\boldsymbol{x}_i) = \mathcal{N}(\mathbf{z}|\boldsymbol{\mu}_i,\boldsymbol{\sigma}_i^2\mathbf{I}),
\end{equation}
\begin{equation}
    p_{\beta^d}(\mathbf{z}) = \mathcal{N}(\mathbf{z}|\boldsymbol{0},\mathbf{I}),
\end{equation}
Since $\mathbf{f}$ is the representative of $\mathbf{x}$, we define the variational generative process based on $\mathbf{f}$ rather than $\mathbf{x}$. Let $e$ be a VAE encoder that maps the image descriptor $\boldsymbol{f}_i$ into the parameters of the $q_{\beta^e}$ distribution $\boldsymbol{\mu}_i$ and $\boldsymbol{\sigma}_i$ for $\boldsymbol{x}_i$. To prevent the interference of the DNN training with the stochastic sampling of $\mathbf{z}$, we utilize the reparameterization trick introduced in~\cite{Kingma:2014} to generate $\boldsymbol{z}_i$ as follows:
\begin{equation}
\boldsymbol{z}_i = \boldsymbol{\mu}_i + \boldsymbol{\sigma}_i \cdot \boldsymbol{\epsilon}_i;\;\boldsymbol{\epsilon}_i\sim \mathcal{N}(\boldsymbol{0},\mathbf{I}).
\end{equation}
Let $t^g : \beta^t, \mathcal{Z} \mapsto \hat{\mathcal{Y}}^g$ be a generative task head that maps the latent into the corresponding label prediction associated with the image $\boldsymbol{x}_i$ [i.e., $t^g(\boldsymbol{z}_i;\beta^t) = \hat{\boldsymbol{y}}_i^g$], where $\beta^t$ is the task head parameters and $\hat{\mathcal{Y}}^g$ is the set of all predicted image labels. $t^g$ is chosen as the duplicate of $t^d$, but they are associated to different model parameters (i.e., $\beta^t \neq \gamma$). Let $r : \beta^r, \mathcal{Z} \mapsto \hat{\mathcal{F}}$ be a feature decoder that maps the latent into the reconstructed image descriptor $\hat{\boldsymbol{f}}_i$ for $\boldsymbol{x}_i$, where $\beta^r$ is the feature decoder parameters. $t^g$ models $p_{\beta^d}(\mathbf{y}|\mathbf{z})$, while $r$ models $p_{\beta^d}(\mathbf{x}|\mathbf{z})$. Accordingly, the generative task head $t^g$ and the feature decoder $r$ both form the VAE decoder (i.e., $\beta^d = \beta^t \cup \beta^r$).

To accurately model $p(\mathbf{x},\mathbf{y})$, the VAE parameters $\beta = \beta^e \cup \beta^d$ can be learned by maximizing the ELBO defined in (3). To this end, we define the error functions for: i) the first term of the ELBO based on mean squared error loss function $\mathcal{L}_{\text{MSE}}$; ii) second term of the ELBO based on the loss function $\mathcal{L}$ considered for discriminative reasoning; and iii) third term of the ELBO based on the known functional forms of $q_{\beta^e}(\mathbf{z}|\boldsymbol{x}_i)$ and $p_{\beta^d}(\mathbf{z})$. Accordingly, the objective function $O_g$ associated with generative reasoning for a set of samples $\mathcal{S}$ is written as follows:
\begin{equation}
\begin{aligned}
    O_g(\mathcal{S}) &= \frac{1}{|\mathcal{S}|} \sum_{(\boldsymbol{x}_i,\boldsymbol{y}_i)\in\mathcal{S}}\mathcal{L}_{\text{MSE}}(r(\boldsymbol{z}_i;\beta^r),\boldsymbol{f}_i) \\
     &+\frac{1}{|\mathcal{S}|}\sum_{(\boldsymbol{x}_i,\boldsymbol{y}_i)\in\mathcal{S}}\mathcal{L}(t^g(\boldsymbol{z}_i;\beta^t),\boldsymbol{y}_i)\\
     &+\frac{1}{2}\sum_{j=1}^J \big(1+\text{log}(\boldsymbol{\sigma}_{i,j}^2)-\boldsymbol{\mu}_{i,j}^2-\boldsymbol{\sigma}_{i,j}^2\big),
\end{aligned}
\end{equation}
where $\boldsymbol{\mu}_{i,j}$ and $\boldsymbol{\sigma}_{i,j}$ are the $j$th element of the vectors $\boldsymbol{\mu}_{i}$ and $\boldsymbol{\sigma}_{i}$, respectively, while $J$ is their length. We would like to note that minimizing this objective function leads to maximizing the ELBO, and thus more accurate modelling of $p(\mathbf{x},\mathbf{y})$. For the derivation of the KL divergence term in the ELBO, the reader is referred to~\cite{Kingma:2014}.

It is worth noting that the proposed integration of generative reasoning into discriminative reasoning does not depend on the selection of the loss function $\mathcal{L}$ and discriminative task head, and thus can be applied to most of the supervised DL-based IRL methods. It also does not require an additional CNN backbone as image encoder, since we define the variational generative process based on $\mathbf{f}$. Thus, the considered VAE is directly branched out from the CNN backbone to learn image representations based on generative and discriminative reasoning together.

\subsection{Label Noise Robust Hybrid Representation Learning}
The proposed GRID approach aims to jointly model the posterior and joint distributions of annotated images in a single learning procedure, while achieving label noise robust IRL. To this end, we introduce a label noise robust hybrid representation learning strategy to model images through: i) generative reasoning for the training samples with noisy labels; and ii) discriminative reasoning for the remaining samples in the training data. For the sake of simplicity, we refer training samples with noisy labels as noisy samples, and those with correct labels as clean samples hereafter. It is noted that generative reasoning is less annotation dependent compared to discriminative reasoning due to modelling the joint distribution $p(\mathbf{x},\mathbf{y})$ in a probabilistic generative process. Thus, for discriminative reasoning, the loss value differences between noisy samples and clean samples are higher compared to generative reasoning. The proposed integration of generative reasoning into discriminative reasoning allows to automatically detect noisy samples based on the loss values of $\mathcal{L}$ incurred through generative and discriminative reasoning. Accordingly, we decide whether a training sample is noisy or clean based on the loss values acquired from discriminative and generative task heads. To this end, we define our automatic noisy sample detection procedure as follows. A training sample is considered as noisy if it leads to a significantly higher loss value from the discriminative task head compared to the generative task head. For a given mini-batch $\mathcal{B}$, we first sort the differences of normalized loss values acquired from discriminative and generative task heads. This can be defined as a non-decreasing sequence $A$ as follows:
\begin{equation}
\begin{aligned}
    &A=(a_k)_{k=1}^{|\mathcal{B}|},\;a_k\geq a_{k+1} \\
    &a_k \in \{\mathcal{L}(\hat{\boldsymbol{y}}_i^d, \boldsymbol{y}_i)-\mathcal{L}(\hat{\boldsymbol{y}}_i^g, \boldsymbol{y}_i)\}_{(\boldsymbol{x}_i,\boldsymbol{y}_i) \in \mathcal{B}}\;\forall k,\\
\end{aligned}
\end{equation}
where loss values are normalized based on the min-max scaling strategy. Then, we divide $\mathcal{B}$ into the set $\mathcal{W}$ of noisy samples and the set $\mathcal{C}$ of clean samples, where $\mathcal{B} = \mathcal{W} \cup \mathcal{C};\;\mathcal{W} \cap \mathcal{C} = \emptyset$, as follows:
\begin{equation}
\mathcal{W} = \{(\boldsymbol{x}_i,\boldsymbol{y}_i)|(\boldsymbol{x}_i,\boldsymbol{y}_i)\!\in\!\mathcal{B} \land \mathcal{L}(\hat{\boldsymbol{y}}_i^d, \boldsymbol{y}_i)\!-\!\mathcal{L}(\hat{\boldsymbol{y}}_i^g, \boldsymbol{y}_i)\!>\! a_{\lambda} \}
\end{equation}
\begin{equation}
    \mathcal{C} = \{(\boldsymbol{x}_i,\boldsymbol{y}_i)| (\boldsymbol{x}_i,\boldsymbol{y}_i)\!\in\!\mathcal{B} \land \mathcal{L}(\hat{\boldsymbol{y}}_i^d, \boldsymbol{y}_i)\!-\!\mathcal{L}(\hat{\boldsymbol{y}}_i^g, \boldsymbol{y}_i)\! \leq\! a_{\lambda}\}.
\end{equation}
$\mathcal{W}$ includes the samples from $\mathcal{B}$ associated with the $\lambda \in \{1,2,\ldots,|\mathcal{B}|\}$ largest elements of $A$ (i.e., the $\lambda$ highest loss value differences), while $\mathcal{C}$ includes the rest of the samples from $\mathcal{B}$. $\lambda$ is a hyper-parameter of the proposed GRID approach. 

To learn the model parameters associated with discriminative and generative reasoning, one could directly apply optimization to jointly minimize $O_g(\mathcal{B})$ and $O_d(\mathcal{B})$. This leads to optimization of the objectives for all samples in $\mathcal{B}$ based on both generative and discriminative reasoning. When it is applied to the parameters $\theta$ of the CNN backbone $\phi$, it can limit to exploit the effectiveness of generative reasoning for noisy samples and that of discriminative reasoning for clean samples due to interference of different learning characteristics. Accordingly, the model parameters $\theta$ are updated based on whether a sample is assigned to $\mathcal{W}$ or $\mathcal{C}$. Accordingly, the update rule for $\theta$ is written as follows:
\begin{equation}
    \theta \leftarrow \theta - \eta \nabla_{\theta}\Big(\frac{|\mathcal{W}|O_g(\mathcal{W}) + |\mathcal{C}|O_d(\mathcal{C})}{|\mathcal{B}|}\Big),
\end{equation}
where $\eta$ is the learning rate. It is noted that we define the variational generative process based on the image descriptors. Accordingly, for the backbone parameters, the first and the third terms of the ELBO is assumed to be 0 (see (\ref{eq:elbo2})). Then, the update rule can be written based on only $\mathcal{L}$ as follows:
\begin{equation}
    \theta \leftarrow \theta - \frac{\eta\nabla_{\theta}}{|\mathcal{B}|}\;\Big(\sum_{\mathclap{\;\;\;\;\;(\boldsymbol{x}_i,\boldsymbol{y}_i)\in\mathcal{W}}}\mathcal{L}(\hat{\boldsymbol{y}}_i^g, \boldsymbol{y}_i) + \sum_{\mathclap{\;\;\;\;(\boldsymbol{x}_i,\boldsymbol{y}_i)\in\mathcal{C}}}\mathcal{L}(\hat{\boldsymbol{y}}_i^d, \boldsymbol{y}_i)\Big).
\end{equation}
Based on this update rule, the CNN backbone parameters are updated only to minimize $\mathcal{L}$, whose values are obtained from generative task head for noisy samples and discriminative task head for clean samples. Accordingly, image representations are learned based on: i) the generative reasoning for noisy training samples; and ii) discriminative reasoning for clean samples. However, for the remaining model parameters, it is important to maintain the characteristics of discriminative and generative reasoning throughout the training. Accordingly, discriminative task head parameters $\gamma$ are updated based on $O_d(\mathcal{B})$, while the VAE parameters $\beta$ are updated based on $O_g(\mathcal{B})$ as follows:
\begin{equation}
\begin{aligned}
    \gamma \leftarrow & \gamma - \eta\nabla_{\gamma}O_d(\mathcal{B})\\
    \beta \leftarrow & \beta - \eta\nabla_{\beta}O_g(\mathcal{B}).
\end{aligned}
\end{equation}

Due to the automatic detection of noisy samples and learning image representation space (characterized by the CNN backbone) accordingly, the proposed GRID approach leverages the effectiveness of both discriminative and generative reasoning. This leads to learning image representations robust to label noise without overfitting on noisy labels as in discriminative learning. It is worth mentioning that the proposed GRID approach is independent from the DNN architecture, loss function $\mathcal{L}$, learning task, annotation type being considered and the type of label noise present in training data. In this paper, we assess our approach under two scenarios, where training images are annotated with: 1) scene-level noisy multi-labels; and 2) pixel-level noisy labels. Under these scenarios, we consider three learning tasks with the corresponding loss functions and DNN architectures (see Section~\ref{exp_setup} for the details).

\section{Data Set Description and Experimental Setup}
\subsection{Data Set Description}
We conducted experiments on the BigEarthNet-S2~\cite{BigEarthNet-S2} and the DLRSD~\cite{DLRSD} benchmark image archives. We employed a subset of BigEarthNet-S2 that includes 14,832 Sentinel-2 satellite multispectral images acquired over Serbia in summer. Each image is a section of: i) 120$\times$120 pixels for 10m bands; ii) 60$\times$60 pixels for 20m bands; and iii) 20$\times$20 pixels for 60m bands. It is noted that bicubic interpolation is applied to 20m bands, while 60m bands are excluded from the experiments. Each image was annotated with multi-labels from the CORINE Land Cover Map (CLC) database of the year 2018. For the experiments, we utilized the 19 class nomenclature presented in~\cite{BigEarthNet-S2}. We also extracted the CLC land cover map of each image for the selection of $\mathcal{L}$ (which requires the availability of land-cover maps during training). The DLRSD archive includes 2,100 aerial images. Each image has the size of 256 $\times$ 256 pixels with a spatial resolution of 30 cm, and annotated with both multi-labels and pixel-level labels, for which the class nomenclature is defined in~\cite{Chaudhuri:2018}. For the experiments, BigEarthNet-S2 was divided into training, validation and test sets with the ratios of 52\%, 24\% and 24\%, respectively. DLRSD was divided into training, validation and test sets with the ratios of 70\%, 10\%, 20\%, respectively.

\subsection{Experimental Setup}
\label{exp_setup}
To conduct experiments, we considered two different scenarios, where training images are annotated with: 1) scene-level noisy multi-labels; and 2) pixel-level noisy labels. For these scenarios, we tested our approach under three learning tasks with their corresponding loss functions and DNN architectures that are explained in detail in the following.

In the first scenario, IRL is achieved based on supervised multi-label image classification. For this scenario, binary cross entropy (BCE) loss function $\mathcal{L}_{BCE}$ was chosen as $\mathcal{L}$ of the proposed GRID approach. Accordingly, each of the generative and discriminative task heads includes an FC layer as a classifier that produces multi-label class probabilities. The proposed approach applied to this scenario is denoted as \gridcls~hereafter. 

In the second scenario, IRL is achieved by: 1) semantic segmentation for land cover map generation based on pixel-wise cross entropy loss function (denoted as \gridseg~hereafter); and 2) multi-label co-occurrence prediction based on region representation learning (RRL) loss function $\mathcal{L}_{RRL}$ introduced in~\cite{Sumbul:2021:2} (denoted as \gridgt~hereafter). For \gridseg, each of the generative and discriminative task heads consists of three transposed convolutional layers with the filters of 64, 32 and the number of considered classes. For \gridgt, each of the generative and discriminative task heads includes an FC layer that produces the prediction for graph driven region-based representations. The reader is referred to~\cite{Sumbul:2021:2} for the details. 

For both scenarios, we employed the DenseNet-121 architecture~\cite{Huang:2017} as the CNN backbone, and utilized the latent dimension of 128 for the VAE encoder. The feature decoder of VAE employs an FC layer with the hidden unit size of image descriptor dimension (which is 1024 for DenseNet-121) for \gridcls~and \gridgt, while the FC layer is replaced with a convolutional layer with the kernel size of 1$\times$1 for \gridseg. The parameter $\lambda = k|\mathcal{B}|/100$ was varied as $k \in \{0,10,\ldots,90\}$ when $k$ shows the percentage of each mini-batch that is identified as the set of noisy samples (denoted as $\lambda_{k\%}\;\forall k$ hereafter).

To assess the robustness of our approach to label noise for both scenarios, we applied synthetic label noise injection to the training sets in the range of $[10\%,60\%]$ with the step size of $10\%$. In particular, for scene-level annotations, the set of class labels are randomly chosen from the training label set $\mathcal{Y}$ based on a synthetic label noise injection ratio (SLNIR). Then, each selected class label is randomly changed into one of other class labels, which are not associated with the corresponding image. This ensures that both missing and wrong labels are considered as noisy annotations. For pixel-level annotations, $\mathcal{Y}$ is converted to the set of unique class labels associated with each image prior to random selection based on SLNIR. Then, changed classes are reflected to all relevant pixel labels.

\begin{figure*}[t]
    \input{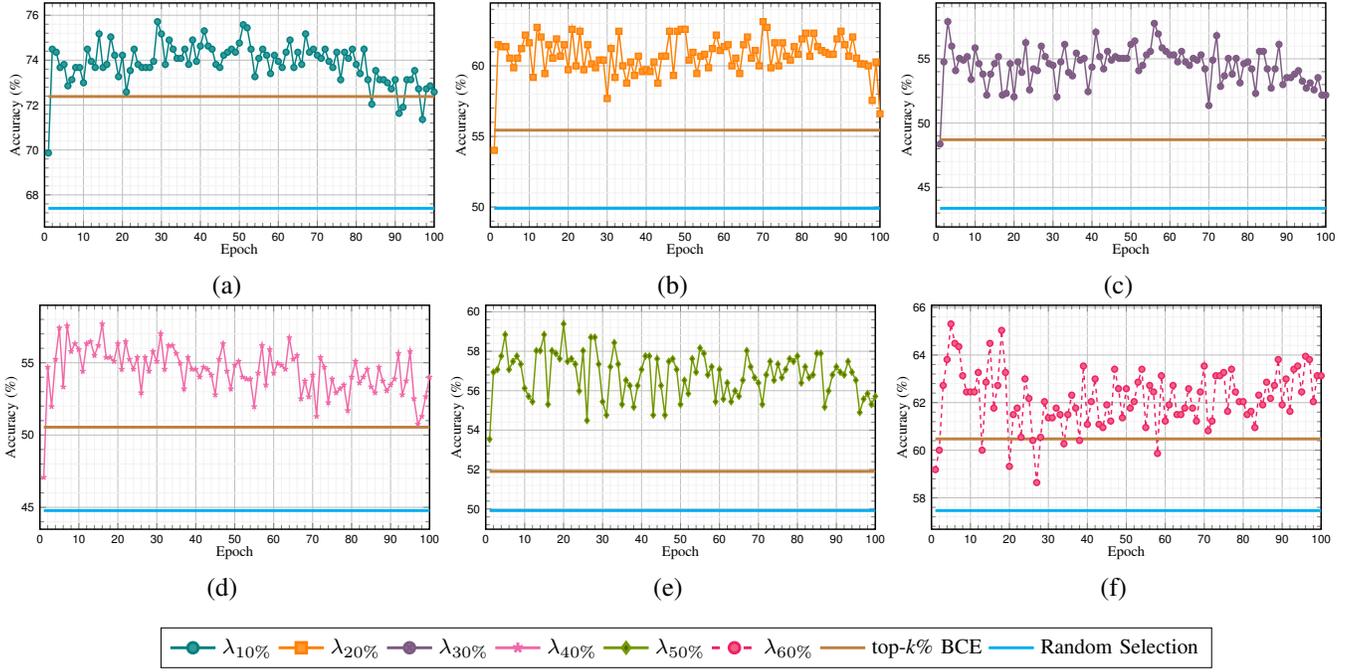}
    \caption{Noisy sample detection accuracy of the proposed \gridcls~approach versus epoch compared to selecting noisy samples: i) based on the top-\textit{k}\% BCE values; and ii) randomly when SLNIR is (a) 10\%, (b) 20\%, (c) 30\%, (d) 40\%, (e) 50\%, (f) 60\%; and $k$ for $\lambda_{k\%}$ and top-\textit{k}\% is set as equal to the SLNIR value (DLRSD archive).}    
    \label{fig:noise_detection_accuracy_DLRSD_NR-Cls-DRL-v3}
\end{figure*}
We conducted experiments related to all scenarios and learning tasks on a single application for the sake of simplicity. This application was selected as content-based image retrieval (CBIR), for which accurately learning image features is of great importance. To apply CBIR after learning image representations, for each archive, we employed the training set for selecting query images, while images were retrieved from the test set. We performed the hyper-parameter selection of our approach on the validation set in the context of CBIR. After randomly initializing the model parameters, we trained our approach for 100 epochs by using the Adam gradient descent optimization algorithm with the initial learning rate of $10^{-3}$ and the mini-batch size of $128$. After IRL is achieved by the proposed approach, we obtained the features of query and archive images from the last layer of the backbone. Then, to apply CBIR, similarity matching of these features was performed by using the $\chi^2$-distance measure. CBIR results are provided in terms of normalized discounted cumulative gains (NDCG), which was averaged on the 20 and 30 most similar images for DLRSD and BigEarthNet-S2, respectively.

For the two above-mentioned scenarios, we carried out experiments to: 1) perform a sensitivity analysis; 2) conduct an ablation study; and 3) compare our approach with the state-of-the-art methods in the framework of CBIR. Under the first scenario, we compared our \gridcls~approach with: 1) the early-learning regularization (ELR) framework~\cite{Liu:2020}; 2) the joint training with co-regularization (JoCoR) approach~\cite{Wei:2020}; IRL with multi-label classification by using 3) focal loss (denoted as FL)~\cite{Lin:2020}; 4) asymmetric loss (denoted as ASL)~\cite{Ridnik:2021}; and 5) the standard binary cross entropy (BCE) loss. It is worth noting that ELR, JoCoR and FL are originally introduced for single-label classification problems. By following~\cite{Burgert:2022}, we adapted them to multi-label classification. Under the second scenario, we compared our \gridseg~approach with: 1) the high-resolution land cover mapping through learning with noise correction method~\cite{Dong:2022} (denoted as LNC); and 2) IRL with semantic segmentation by the standard pixel-wise cross-entropy loss (PCE). For the second scenario, we also compared our \gridgt~approach with IRL with multi-label co-occurrence prediction based on RRL loss~\cite{Sumbul:2021:2} (denoted as RRL). For each comparison with our approach, we used the same CNN backbone and the same task heads. 

\section{Experimental Results}
\subsection{Sensitivity Analysis of the Proposed Approach}
In this sub-section, we present the results of the sensitivity analysis of the proposed approach under scene-level noisy labels (i.e., first scenario) and pixel-level noisy labels (i.e., second scenario) in terms of different values of the $\lambda$ hyper-parameter at different values of SLNIR. For both scenarios, we also assessed the effectiveness of our automatic noisy sample detection procedure in terms of the noisy sample detection accuracy, while it is compared to selecting noisy samples: i) based on the top-\textit{k}\% loss values of the trained BCE (denoted as top-\textit{k}\% (BCE)) and PCE (denoted as top-\textit{k}\% (PCE)) models; and ii) randomly.

\begin{table}[t] %!htbp
\caption{Results (\%) Obtained by the Proposed \gridcls~Approach for Different Values of $\lambda$ and SLNIR (\%) (DLRSD archive)}
\label{table:alpha_comp_table_DLRSD_NR-Cls-DRL-v2}
\begin{table}[t] %!htbp
\caption{Results (\%) Obtained by the Proposed \gridcls~Approach for Different Values of $\lambda$ and SLNIR (\%) (DLRSD archive)}
\label{table:alpha_comp_table_DLRSD_NR-Cls-DRL-v2}
\inputtable{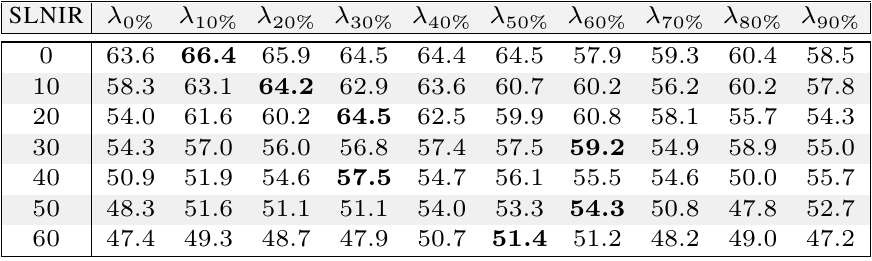}
\end{table}
\end{table}
\begin{figure*}[t]
    \input{sources/plots/noise_detection_accuracy_BEN_NR-Cls-DRL-v3}
    \caption{Noisy sample detection accuracy of the proposed \gridcls~approach versus epoch compared to selecting noisy samples: i) based on the top-\textit{k}\% BCE values; and ii) randomly when SLNIR is (a) 10\%, (b) 20\%, (c) 30\%, (d) 40\%, (e) 50\%, (f) 60\%; and $k$ for $\lambda_{k\%}$ and top-\textit{k}\% is set as equal to the SLNIR value (BigEarthNet-S2 archive).}
    \label{fig:noise_detection_accuracy_BEN_NR-Cls-DRL-v3}
\end{figure*}
\begin{table}[t] %!htbp
\caption{Results (\%) Obtained by the Proposed \gridcls~Approach for Different Values of $\lambda$ and SLNIR (\%) (BigEarthNet-S2 archive)}
\label{table:alpha_comp_table_BEN_NR-Cls-DRL-v2}
\begin{table}[t] %!htbp
\caption{Results (\%) Obtained by the Proposed \gridcls~Approach for Different Values of $\lambda$ and SLNIR (\%) (BigEarthNet-S2 archive)}
\label{table:alpha_comp_table_BEN_NR-Cls-DRL-v2}
\inputfig{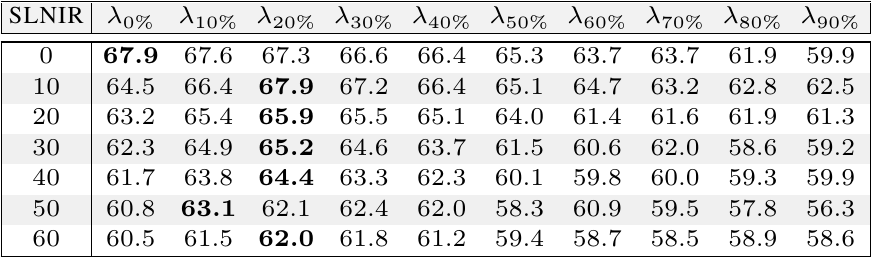}
\end{table}
\end{table}
\subsubsection{First Scenario (Scene-Level Noisy Labels)}
Tables~\ref{table:alpha_comp_table_DLRSD_NR-Cls-DRL-v2} and~\ref{table:alpha_comp_table_BEN_NR-Cls-DRL-v2} show the results of \gridcls~for the DLRSD and BigEarthNet-S2 archives, respectively. One can see from the Table{~\ref{table:alpha_comp_table_DLRSD_NR-Cls-DRL-v2}} that when the level of training label noise increases, our approach achieves generally higher scores by detecting more training samples as noisy with higher values of $\lambda$ for the DLRSD archive. However, as it can be seen from Table{~\ref{table:alpha_comp_table_BEN_NR-Cls-DRL-v2}}, our \gridcls~approach achieves the highest scores when 20\% of each mini-batch is identified as noisy ($\lambda_{20\%}$) for most of the SLNIR values for BigEarthNet-S2. It is worth noting that BigEarthNet-S2 includes a higher number of images compared to DLRSD, and thus there is a lower risk of overfitting to noisy labels. Accordingly, when a training set size is higher than a certain extent as in BigEarthNet-S2, our approach is capable of achieving a high performance with lower values of $\lambda$ under even a high label noise rate. This is due to the fact that each noisy sample affects differently the over-fitting of DNN parameters on noisy labels. It is worth noting that our noise sample detection procedure selects training samples, which lead to the highest differences of loss values acquired from discriminative and generative task heads, as noisy samples. Thus, it selects the training samples as noisy, which most affects the interference of noisy labels compared to the remaining noisy samples. However, when the rate of label noise in a training set is high for a small dataset like DLRSD, our approach requires to increase the effect of generative reasoning through detecting a higher number of noisy samples (i.e., a high value of $\lambda$) for more accurate IRL. In greater details, when no training sample is detected as noisy ($\lambda_{0\%}$), our approach achieves lower results compared to $0 < k < 50$ for $\lambda_{k\%}$ under almost all SLNIR values. This is due to the fact that when $k=0$, IRL is achieved through only discriminative reasoning that increases the risk of overfitting to noisy labels. This shows the importance of detecting noisy samples and employing generative reasoning for IRL on those samples. By considering that there is not a single $\lambda$ value that provides the highest scores under all SLNIR values for DLRSD, we set it based on the results on BigEarthNet-S2. Accordingly, for the rest of the experiments, we set $\lambda$ of \gridcls~to $\lambda_{20\%}$.

\begin{figure*}[t]
    \input{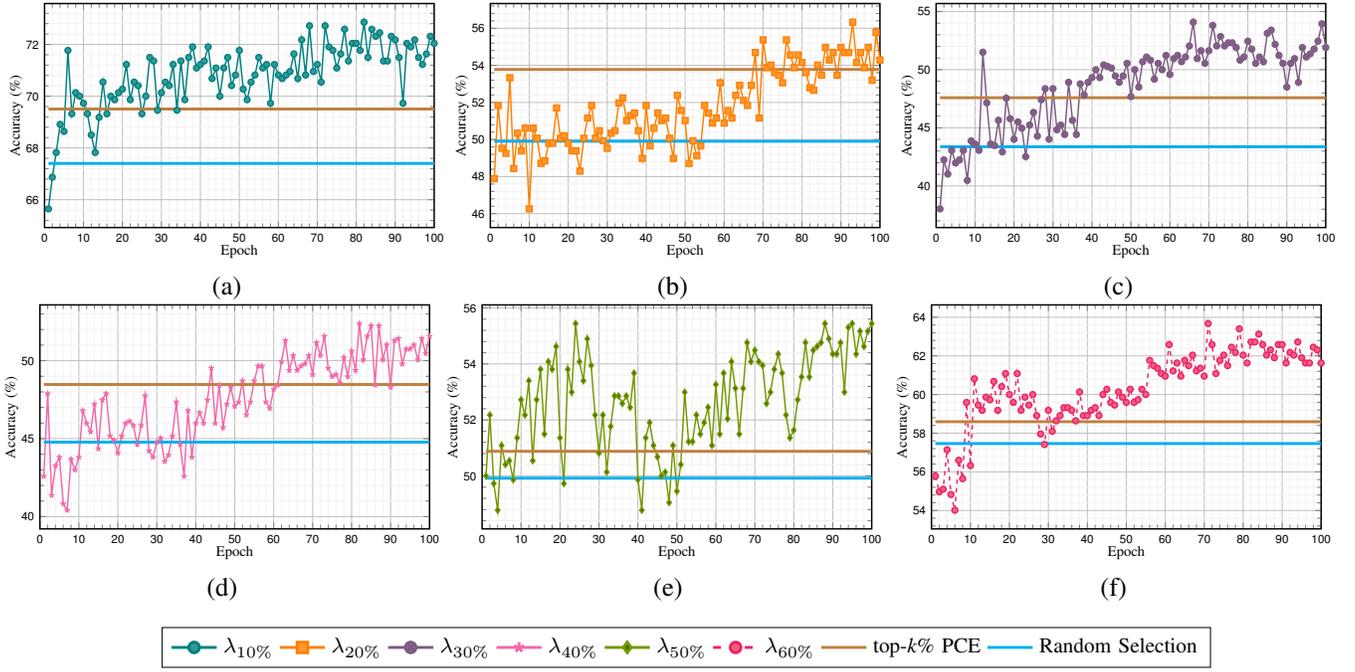}
    \caption{Noisy sample detection accuracy of the proposed \gridseg~approach versus epoch compared to selecting noisy samples: i) based on the top-\textit{k}\% PCE values; and ii) randomly when SLNIR is (a) 10\%, (b) 20\%, (c) 30\%, (d) 40\%, (e) 50\%, (f) 60\%; and $k$ for $\lambda_{k\%}$ is set as equal to the SLNIR value (DLRSD archive).}
    \label{fig:noise_detection_accuracy_DLRSD_NR-Seg-DRL-v3}
\end{figure*}
We would like to note that if the value of $\lambda$ is high, there is a risk of detecting training samples with correct labels (i.e., clean samples) as noisy samples. To analyze the effectiveness of our automatic noisy sample detection procedure, Figures{~\ref{fig:noise_detection_accuracy_DLRSD_NR-Cls-DRL-v3}} and{~\ref{fig:noise_detection_accuracy_BEN_NR-Cls-DRL-v3}} show the noisy sample detection accuracies for our procedure,  top-\textit{k}\% (BCE) and random selection when $k$ for $\lambda_{k\%}$ and top-\textit{k}\% is set as equal to the SLNIR value (e.g., $\lambda_{20\%}$ and top-20\% for SLNIR $=20\%$) for DLRSD and BigEarthNet-S2, respectively. One can observe from the figures that our approach detects noisy samples more accurately than top-\textit{k} (BCE) and random selection under almost all SLNIR values. This shows the effectiveness of our automatic noisy sample detection procedure in the proposed approach. It can be also seen from the figures that after a certain number of training epochs, noisy sample detection accuracy starts to decrease for most of the SLNIR values. It is due the fact that as the proposed approach combines generative and discriminative reasoning during training, image representation space encoded by the CNN backbone starts to become robust to noisy samples. Then, for our approach detecting noisy samples becomes harder and harder based on the image features from the backbone as training continues. This leads to decrease in noisy sample detection accuracy after a certain number of epochs. In greater detail, our approach trained on BigEarthNet-S2 provides higher detection accuracy compared to that on DLRSD especially on higher SLNIR values. It is due to the higher number of training samples in BigEarthNet-S2 compared to DLRSD that allows our approach to learn model parameters and to detect noisy samples more accurately. 

\begin{figure*}[ht!]
    \input{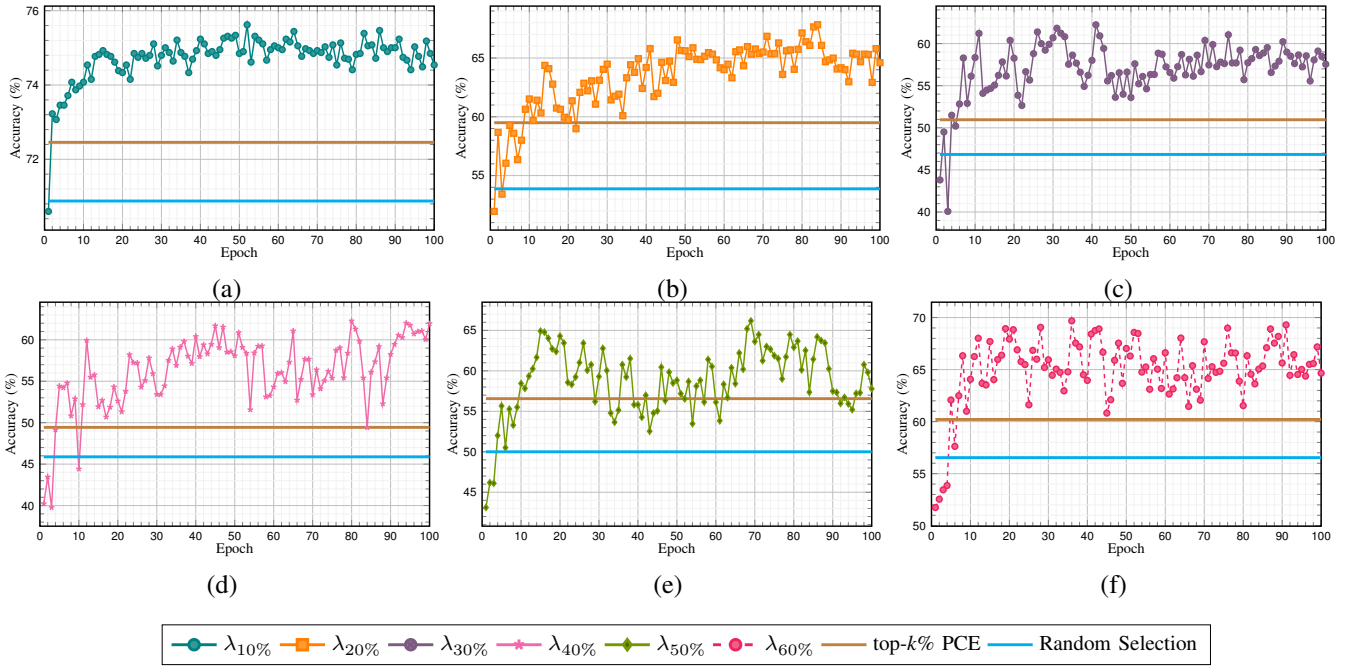}
    \caption{Noisy sample detection accuracy of the proposed \gridseg~approach versus epoch compared to selecting noisy samples: i) based on the top-\textit{k}\% PCE values; and ii) randomly when SLNIR is (a) 10\%, (b) 20\%, (c) 30\%, (d) 40\%, (e) 50\%, (f) 60\%; and $k$ for $\lambda_{k\%}$ is set as equal to the SLNIR value (BigEarthNet-S2 archive).}
    \label{fig:noise_detection_accuracy_BEN_NR-Seg-DRL-v3}
\end{figure*}
\begin{figure*}[ht!]
    \centering
    \input{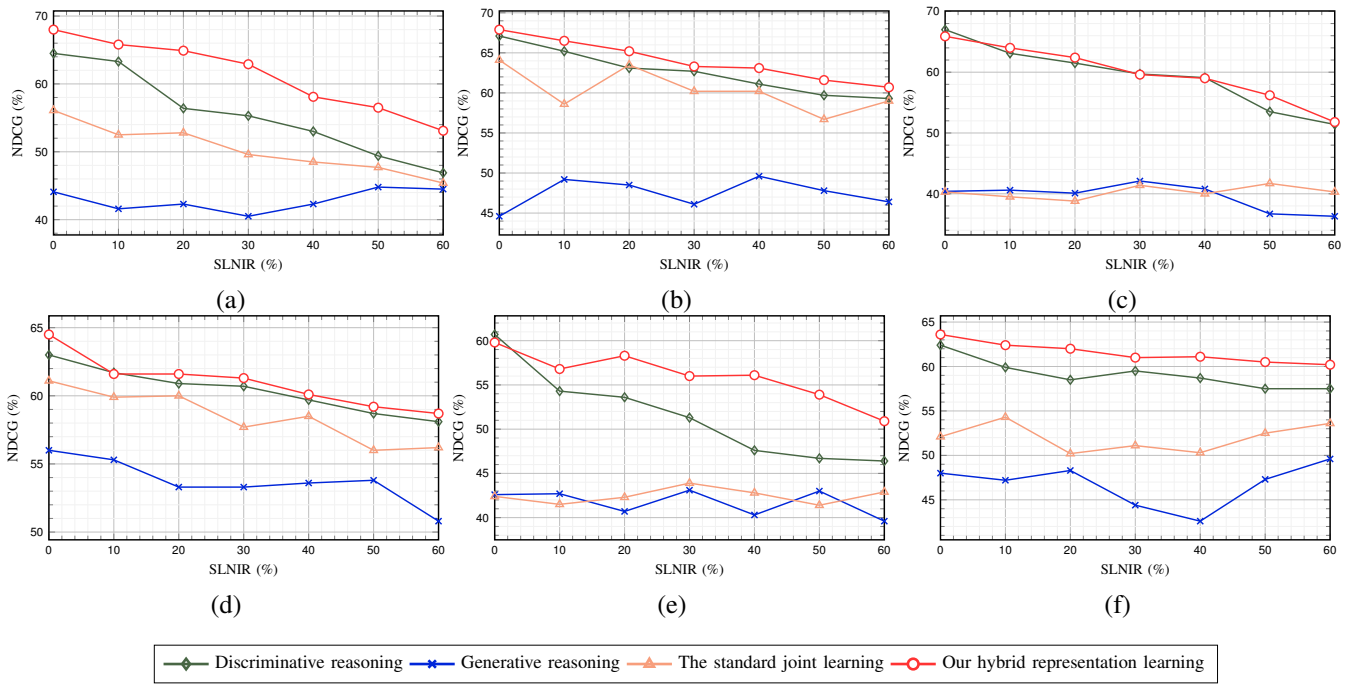}
    \caption{Results obtained by using: 1) discriminative reasoning; 2) generative reasoning; 3) their standard joint learning; and 4) our label noise robust hybrid representation learning strategy for different values of SLNIR when IRL is achieved by: i) multi-label classification on (a) DLRSD and (b) BigEarthNet-S2; ii) semantic segmentation on (c) DLRSD and (d) BigEarthNet-S2; and iii) multi-label co-occurrence prediction on (e) DLRSD and (f) BigEarthNet-S2.} 
    \label{fig:ablation_figure_v3}
\end{figure*}
\begin{table}[t] %!htbp
\caption{Results (\%) Obtained by the Proposed \gridseg~Approach for Different Values of $\lambda$ and SLNIR (\%) (DLRSD archive)}
\label{table:alpha_comp_table_DLRSD_NR-Seg-DRL-v2}
\begin{table}[t] %!htbp
\caption{Results (\%) Obtained by the Proposed \gridseg~Approach for Different Values of $\lambda$ and SLNIR (\%) (DLRSD archive)}
\label{table:alpha_comp_table_DLRSD_NR-Seg-DRL-v2}
\inputfig{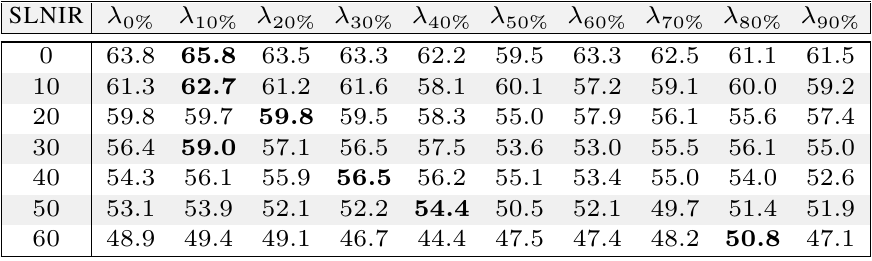}
\end{table}
\end{table}
\begin{table}[t] %!htbp
\caption{Results (\%) Obtained by the Proposed \gridseg~Approach for Different Values of $\lambda$ and SLNIR (\%) (BigEarthNet-S2 archive)}
\label{table:alpha_comp_table_BEN_NR-Seg-DRL-v2}
\begin{table}[t] %!htbp
\caption{Results (\%) Obtained by the Proposed \gridseg~Approach for Different Values of $\lambda$ and SLNIR (\%) (BigEarthNet-S2 archive)}
\label{table:alpha_comp_table_BEN_NR-Seg-DRL-v2}
\inputfig{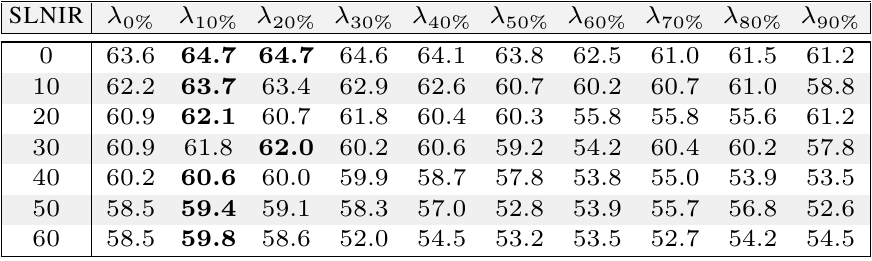}
\end{table}
\end{table}

\subsubsection{Second Scenario (Pixel-Level Noisy Labels)}
Tables~\ref{table:alpha_comp_table_DLRSD_NR-Seg-DRL-v2} and \ref{table:alpha_comp_table_BEN_NR-Seg-DRL-v2} show the results of \gridseg~for the DLRSD and BigEarthNet-S2 archives, respectively. By assessing the table, one can observe that as SLNIR value increases the proposed approach achieves the higher scores with higher values of $\lambda$ for DLRSD. However, for BigEarthNet-S2, the proposed \gridseg~approach achieves the highest scores when $\lambda$ is set to $\lambda_{10\%}$ for most of the SLNIR values. This is inline with our conclusion from the first scenario. In greater detail, for most of the SLNIR values, \gridseg~achieves the higher scores with lower values of $\lambda$ compared to \gridcls~for both archives. This is due to the fact that the semantic segmentation task of the second scenario is more complex than the multi-label image classification task. Accordingly, our \gridseg~approach requires increasing the effect of discriminative reasoning over generative reasoning compared to \gridcls~to overcome the complexity of the semantic segmentation task. This can be achieved by decreasing the value of $\lambda$ as it can be seen from the results. Due to the same reason, when no training sample is detected as noisy ($\lambda_{0\%}$), the performance of \gridseg~is less affected compared to \gridcls~for both archives. For the rest of the experiments, we set $\lambda$ of \gridseg~to $\lambda_{10\%}$ based on the BigEarthNet-S2 results similar to the first scenario.

Figures~\ref{fig:noise_detection_accuracy_DLRSD_NR-Seg-DRL-v3} and~\ref{fig:noise_detection_accuracy_BEN_NR-Seg-DRL-v3} show the noisy sample detection accuracies of the proposed \gridseg~approach, top-\textit{k}\% (PCE) and random selection for DLRSD and BigEarthNet-S2, respectively, when $k$ for $\lambda_{k\%}$ and top-\textit{k}\% is set as equal to the SLNIR value. One can see from the figures that \gridseg~is capable of detecting noisy samples with higher accuracy than top-\textit{k} (PCE) and random sampling under most of the SLNIR values for both archives. In particular, the proposed approach achieves higher detection accuracy on BigEarthNet-S2 than DLRSD. These follow our conclusion from the first scenario. This shows that our approach is capable of accurately detecting noisy samples independently from the considered loss function, learning task, DNN and training sample annotation type. In greater detail, unlike the first scenario, after a certain number of training epochs noisy sample detection accuracy of \gridseg~becomes non-decreasing for some SLNIR values. It is due to the relative complexity of semantic segmentation task compared to multi-label image classification task that may require more training epochs for our approach under especially high SLNIR values. Since label noise rate of a training set is assumed to be unknown for our approach, we avoided over-parameterization of hyper-parameters such as number of training epochs. It is noted that the results for the sensitivity analysis of the second scenario were also confirmed through experiments for our \gridgt~approach on both archives (not reported for space constraints).

\begin{figure*}[t]
    \centering
    \input{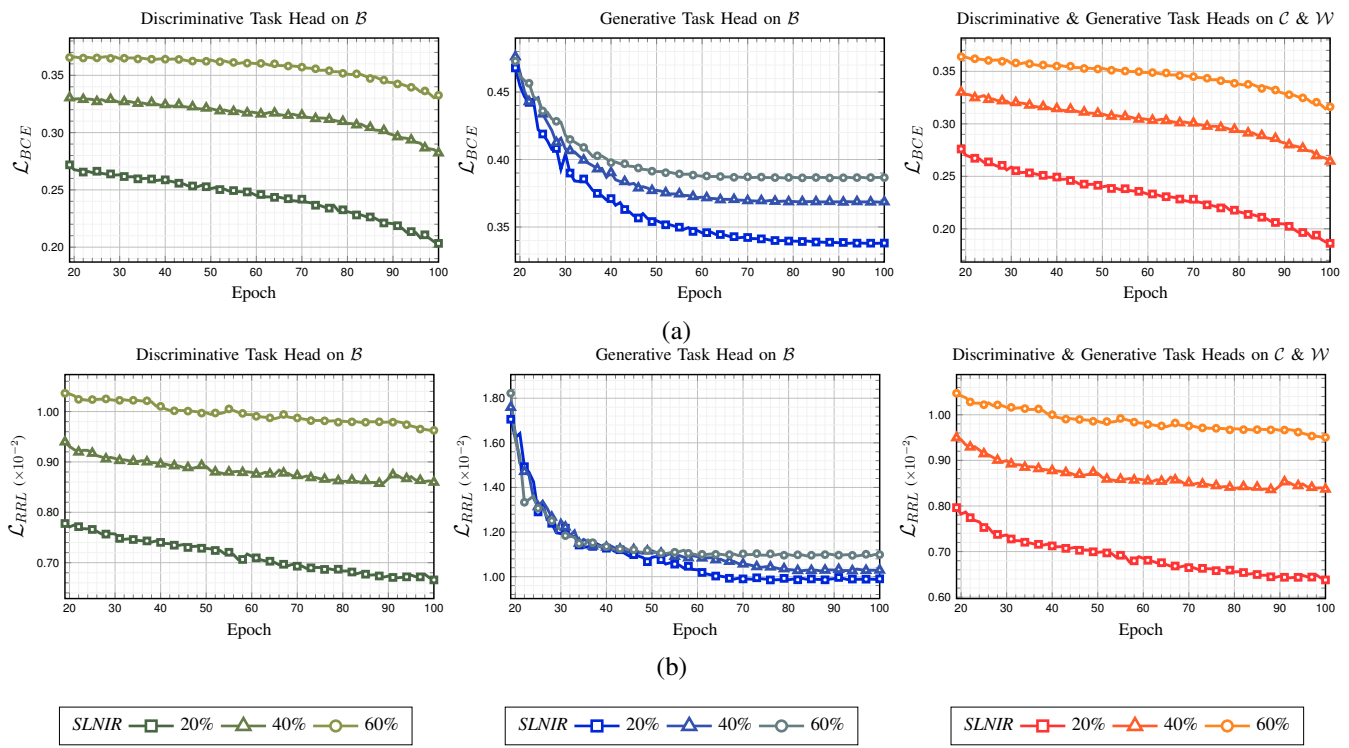}
    \caption{Average loss values versus epoch obtained by: i) the discriminative task head on all the samples; ii) the generative task head on all the samples; and iii) the discriminative and generative task heads on the clean and noisy samples, respectively, during the training of the proposed GRID approach when IRL is achieved by: (a) multi-label classification; and (b) multi-label co-occurrence prediction under different SLNIR values (BigEarthNet-S2 archive).}      
    \label{fig:loss_comp_figure}
\end{figure*}
\subsection{Ablation Study of the Proposed Approach}
In this sub-section, we present an ablation study of our approach to analyze the effectiveness of our label noise robust hybrid representation learning compared to using: i) only discriminative reasoning; ii) only generative reasoning; and iii) their standard joint learning under both first and second scenarios. For the standard joint learning of discriminative and generative reasoning, we jointly minimize $O_g$ and $O_d$ for all the training samples without the detection of noisy samples. This leads to optimization of the all model parameters based on both generative and discriminative reasoning of noisy samples and clean samples together. Fig.~\ref{fig:ablation_figure_v3} shows the results of using: i) discriminative reasoning; ii) generative reasoning; iii) the standard joint learning of discriminative and generative reasoning; and iv) our label noise robust hybrid representation learning strategy under different SLNIR values for both archives. By assessing the figure, one can observe that our label noise robust hybrid representation learning strategy provides the highest scores for most of the SLNIR values independently from the considered scenarios. This shows that our approach is capable of: i) accurately combining generative and discriminative reasoning independently from the considered loss function, learning task and type of annotation; and ii) effectively adjusting the whole learning procedure accordingly for label noise robust IRL. In greater detail, generative reasoning achieves the lowest scores under most of the SLNIR values and considered scenario compared to discriminative reasoning. However, its performance is less affected by the increase in label noise rate compared to discriminative reasoning. This shows the capability of generative reasoning to allow robust learning of image representations under label noise. One can see from the figure that the standard joint learning provides lower scores compared to using only discriminative reasoning for most of the SLNIR values under both scenarios. Learning image representations based on discriminative and generative reasoning on all the training samples may not be accurately achieved due to interference of different learning characteristics. However, when the complementary characteristics of discriminative and generative reasoning is modeled based on our hybrid representation learning strategy, the proposed approach is capable of overcoming this limitation. This is achieved by learning image representations through: i) generative reasoning for the noisy samples based on the loss values obtained from the generative task head; and ii) discriminative reasoning for the clean samples based on the loss values obtained from the discriminative task head. Fig.{~\ref{fig:loss_comp_figure}} shows the average loss values versus epoch during the training of the proposed GRID approach on the BigEarthNet-S2 archive for both scenarios under different SLNIR values. By analyzing the figure, one can observe that the loss values incurred through the generative task head are less affected by the increase in SLNIR values compared to those through the discriminative task head. However, the loss values of the discriminative task head are generally lower compared to those of the generative task head. When the loss values are calculated from the discriminative task head for the clean samples and the generative task head for the noisy samples in our hybrid strategy, the loss values are even further decreased compared to the discriminative task head. These are inline with our conlusions from Fig.{~\ref{fig:ablation_figure_v3}} that shows the importance of the label noise robust hybrid representation learning strategy of the proposed approach.

\begin{figure}[t]
    \centering
    \input{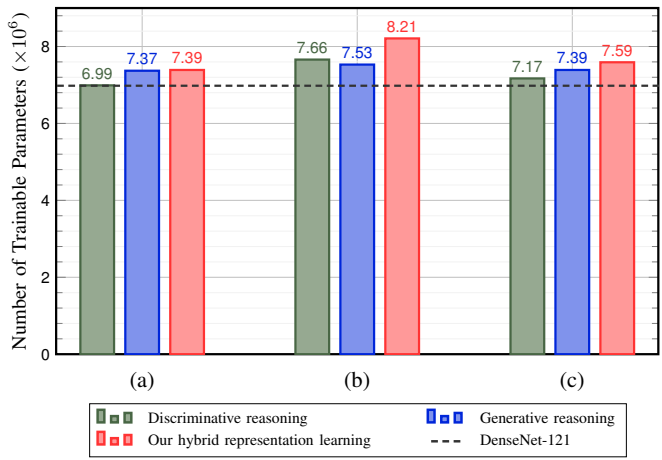}
    \caption{Number of trainable parameters associated to using: 1) only discriminative reasoning; 2) only generative reasoning; 3) our label noise robust hybrid representation learning strategy; and 4) the considered CNN backbone (DenseNet-121) when IRL is achieved by: (a) multi-label classification; (b) semantic segmentation; and (c) multi-label co-occurrence prediction.}      
    \label{fig:ablation_nb_param}
\end{figure}

We would like to note that while the use of only discriminative reasoning requires to learn the CNN backbone parameters $\theta$ and discriminative task head parameters $\gamma$ during training, the use of only generative reasoning requires to learn the VAE parameters $\beta$ and $\theta$. However, for our approach $\theta$, $\beta$ and $\gamma$ are required to be learned that increases the total number of model parameters. Fig.{~\ref{fig:ablation_nb_param}} shows the number of trainable parameters associated to using: i) only discriminative reasoning; ii) only generative reasoning; iii) our label noise robust hybrid representation learning strategy; and iv) the considered CNN backbone (DenseNet-121) for all scenarios. By assessing the figure, one can see that at least 85\% of the trainable parameters are associated with the CNN backbone independently from the considered scenario and the type of IRL. This shows that the selection of CNN backbone is the most important factor affecting the number of model parameters. One can also observe from the figure that our approach leads to at most 7\% higher number of parameters compared to using only discriminative reasoning. However, the effect of this increase on the computational complexity of IRL training is negligible if recent computational units are employed for training. In the case of training under a limited computational power, the number of trainable parameters of our approach can be reduced by using smaller dimensions of the VAE latent and the image descriptor of a lightweight CNN backbone.
\begin{table}[t] %!htbp
\renewcommand{\arraystretch}{1.1}
\setlength\tabcolsep{0pt}
\centering
\scriptsize
\caption{Results (\%) Obtained by BCE, ELR~\cite{Liu:2020}, FL~\cite{Lin:2020}, ASL~\cite{Ridnik:2021}, JoCoR~\cite{Wei:2020} and the Proposed \gridcls~Approach Under Different Values of SLNIR (\%) (DLRSD archive)}
\label{table:soa_table_DLRSD_BCE_v3}
\begin{table}[t] %!htbp
\renewcommand{\arraystretch}{1.1}
\setlength\tabcolsep{0pt}
\centering
\scriptsize
\caption{Results (\%) Obtained by BCE, ELR~\cite{Liu:2020}, FL~\cite{Lin:2020}, ASL~\cite{Ridnik:2021}, JoCoR~\cite{Wei:2020} and the Proposed \gridcls~Approach Under Different Values of SLNIR (\%) (DLRSD archive)}
\label{table:soa_table_DLRSD_BCE_v3}
\inputfig{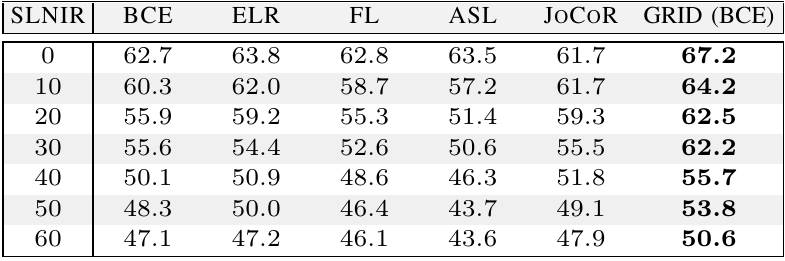}
\end{table}
\end{table}
\begin{table}[t] %!htbp
\renewcommand{\arraystretch}{1.1}
\setlength\tabcolsep{0pt}
\centering
\scriptsize
\caption{Results (\%) Obtained by BCE, ELR~\cite{Liu:2020}, FL~\cite{Lin:2020}, ASL~\cite{Ridnik:2021}, JoCoR~\cite{Wei:2020} and the Proposed \gridcls~Approach Under Different Values of SLNIR (\%) (BigEarthNet-S2 archive)}
\label{table:soa_table_BEN_BCE_v3}
\begin{table}[t] %!htbp
\renewcommand{\arraystretch}{1.1}
\setlength\tabcolsep{0pt}
\centering
\scriptsize
\caption{Results (\%) Obtained by BCE, ELR~\cite{Liu:2020}, FL~\cite{Lin:2020}, ASL~\cite{Ridnik:2021}, JoCoR~\cite{Wei:2020} and the Proposed \gridcls~Approach Under Different Values of SLNIR (\%) (BigEarthNet-S2 archive)}
\label{table:soa_table_BEN_BCE_v3}
\inputfig{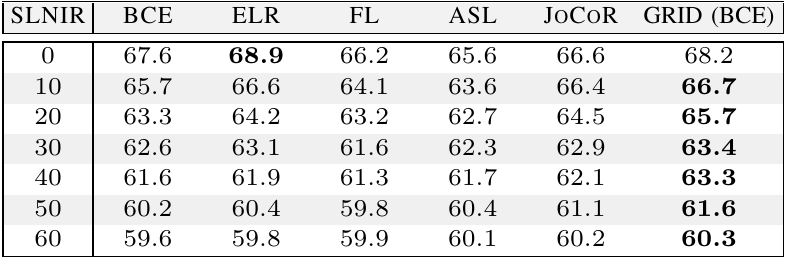}
\end{table}
\end{table}
\begin{table}[t] %!htbp
\setlength\tabcolsep{0pt}
\renewcommand{\arraystretch}{1.1}
\centering
\scriptsize
\caption{Results (\%) Obtained by PCE, LNC~\cite{Dong:2022}, RLL~\cite{Sumbul:2021:2} and the Proposed \gridseg~and \gridgt~Approaches Under Different Values of SLNIR (\%) (DLRSD archive)}
\label{table:soa_table_DLRSD_SEG_v3}
\begin{table}[t] %!htbp
\setlength\tabcolsep{0pt}
\renewcommand{\arraystretch}{1.1}
\centering
\scriptsize
\caption{Results (\%) Obtained by PCE, LNC~\cite{Dong:2022}, RLL~\cite{Sumbul:2021:2} and the Proposed \gridseg~and \gridgt~Approaches Under Different Values of SLNIR (\%) (DLRSD archive)}
\label{table:soa_table_DLRSD_SEG_v3}
\inputfig{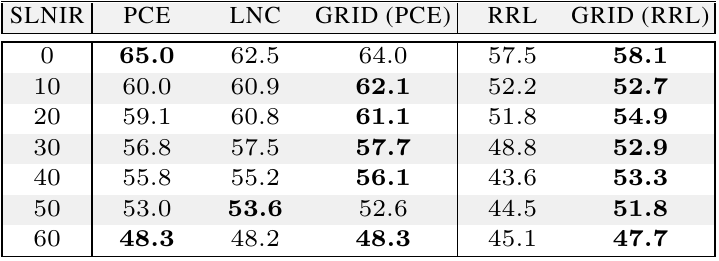}
\end{table}

\end{table}

\begin{table}[t] %!htbp
\renewcommand{\arraystretch}{1.1}
\setlength\tabcolsep{0pt}
\centering
\scriptsize
\caption{Results (\%) Obtained by PCE, LNC~\cite{Dong:2022}, RLL~\cite{Sumbul:2021:2} and the Proposed \gridseg~and \gridgt~Approaches Under Different Values of SLNIR (\%) (BigEarthNet-S2 archive)}
\label{table:soa_table_BEN_SEG_v3}
\begin{table}[t] %!htbp
\renewcommand{\arraystretch}{1.1}
\setlength\tabcolsep{0pt}
\centering
\scriptsize
\caption{Results (\%) Obtained by PCE, LNC~\cite{Dong:2022}, RLL~\cite{Sumbul:2021:2} and the Proposed \gridseg~and \gridgt~Approaches Under Different Values of SLNIR (\%) (BigEarthNet-S2 archive)}
\label{table:soa_table_BEN_SEG_v3}
\inputfig{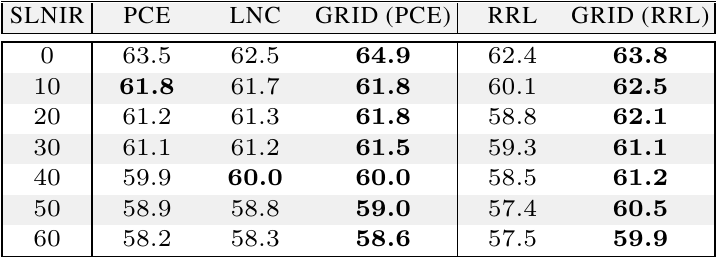}
\end{table}

\end{table}

\subsection{Comparison Among the State-of-the-Art Methods}
In this sub-section, we analyze the effectiveness of the proposed approach compared to different state-of-the-art methods for both scenarios under different values of SLNIR.
\subsubsection{First Scenario (Scene-Level Noisy Labels)}
For the first scenario, we compared our \gridcls~approach with BCE, ELR~\cite{Liu:2020}, FL~\cite{Lin:2020}, ASL~\cite{Ridnik:2021} and JoCoR~\cite{Wei:2020}. Tables~\ref{table:soa_table_DLRSD_BCE_v3}~and \ref{table:soa_table_BEN_BCE_v3} show the corresponding results for DLRSD and BigEarthNet-S2 archives, respectively. By analyzing the tables, one can observe that the proposed \gridcls~approach leads to the highest scores for almost all the SLNIR values on both DLRSD and BigEarthNet-S2 archives. For example, our approach outperforms ELR by almost 8\% NDCG score when SLNIR $=$ 30\% for DLRSD. In detail, it provides more than 3\% higher NDCG score compared to ASL when SLNIR $=$ 10\% for BigEarthNet-S2. As SLNIR value increases, reduction in the NDCG scores is higher for DLRSD compared to BigEarthNet-S2. This is due to the small number of images present in DLRSD that leads to overfitting on noisy labels more easily than BigEarthNet-S2. However, even under high SLNIR values for DLRSD, our approach achieves comparable results with respect to other methods under smaller SLNIR values. As an example, our approach under SLNIR $=$ 40\% achieves similar performance with BCE under SLNIR $=$ 30\%. These results demonstrate the success of the proposed \gridcls~approach compared to other methods when the training images are annotated with scene-level noisy multi-labels. 

\subsubsection{Second Scenario (Pixel-Level Noisy Labels)}
For the second scenario, we compared our \gridseg~approach with PCE and LNC~\cite{Dong:2022}, while \gridgt~was compared with RRL~\cite{Sumbul:2021:2}. Tables~\ref{table:soa_table_DLRSD_SEG_v3}~and \ref{table:soa_table_BEN_SEG_v3} show the corresponding results for DLRSD and BigEarthNet-S2 archives, respectively. One can see from the tables that both \gridseg~and \gridgt~achieve the highest scores compared to other methods under most of the SLNIR values. As an example, when SLNIR $=$ 10\% for DLRSD, \gridseg~achieves more than 1\% higher NDCG score compared to LNC, which is specifically designed for pixel-wise label noise robust semantic segmentation of images. Even when synthetic pixel-wise label noise is not injected to the training sets (SLNIR $=$ 0\%), both \gridseg~and \gridgt~are capable of providing the highest scores for the BigEarthNet-S2 archive. This is due to the fact that even if SLNIR $=$ 0\%, our approach is learning image representations robust to label noise already present in the original training sets. In greater detail, only when SLNIR equals to 50\% and 0\% for DLRSD, our \gridseg~approach is outperformed by LNC and PCE, respectively, with 1\% difference of NDCG scores. However, this is specific to DLRSD archive and not valid for BigEarthNet-S2 archive. These results show the success of our approach compared to other methods when the training samples are annotated with pixel-level noisy labels. This is inline with our conclusion from the first scenario. 

It is worth noting that under two scenarios, we tested our approach with three different loss functions (BCE, PCE and RRL), three different learning tasks (multi-label image classification, semantic segmentation and multi-label co-occurrence prediction) with the corresponding DNN architectures and two different annotation types (scene-level and pixel-level) compared to state-of-the-art methods. The results show that the proposed approach is capable of accurately learning image representations under label noise independently from the considered DNN architecture, loss function, learning task and annotation type. This is due to the capability of our approach to simultaneously leverage the robustness of generative reasoning to noisy labels and the effectiveness of discriminative reasoning for IRL. We would like to also point out that each of the considered state-of-the-art methods is associated to the same number of trainable model parameters with using only discriminative reasoning. Due to this, while our approach achieves IRL under noisy labels more accurately than these methods, it also does not significantly increase the computational complexity of IRL training (see Fig.{~\ref{fig:ablation_nb_param}}).
\section{Conclusion}
In this paper, we have introduced a novel generative reasoning integrated label noise robust deep representation learning (GRID) approach to model the complementary characteristics of discriminative and generative reasoning for IRL under noisy labels. To achieve this, the proposed GRID approach first integrates generative reasoning into discriminative reasoning through a supervised VAE as the probabilistic generative process. Due to this integration, both generative and discriminative reasoning share the same CNN backbone that allows to: 1) model the posterior and joint distributions of annotated images in a single learning procedure; 2) automatically detect training samples with noisy labels based on the loss values acquired from discriminative and generative task heads. This is achieved by the label noise robust hybrid representation learning strategy (which models images through generative reasoning for the training samples with noisy labels and discriminative reasoning for the remaining samples in the training data) in our approach. By this way, the proposed GRID approach learns discriminative image representations through the CNN backbone while preventing interference of noisy labels during training.

It is worth noting that our approach is independent from the type of DNN architecture, loss function, learning task, annotation being considered, label noise present in training data, and can operate with any DL-based IRL method. In addition, GRID does not require the availability of a clean subset of a training set. In this paper, we consider two different scenarios, where training samples are annotated with: 1) scene-level noisy multi-labels; and 2) pixel-level noisy labels. Experimental analysis conducted on two image archives shows the effectiveness of our approach for these scenarios. In particular, the success of our approach is shown under three learning tasks with the corresponding loss functions and DNN architectures at different synthetic label noise injection rates while considering both wrong and missing labels. This shows that the proposed approach accurately learns discriminative image representations, while ensuring the robustness of whole learning procedure towards noisy labels independently from the IRL method being considered. We underline that this is a very important advantage for operational applications that require large training sets (which may include noisy labels).

We would like to point out that our automatic noisy sample detection procedure is controlled by the hyper-parameter $\lambda$. Its selection may be dependent on the level of noisy labels in a training set, which is unknown most of the time in operational scenarios. Accordingly, as a future development of this work, we plan to investigate the strategies for automatically detecting level of noise in training data, and then integrating it into our automatic noisy sample detection procedure for the proposed approach. 

\section*{Acknowledgment}
This work is supported by the European Research Council (ERC) through the ERC-2017-STG BigEarth Project under Grant 759764, and by the German Research Foundation through the IDEAL-VGI project under Grant 424966858, and by the European Space Agency (ESA) through the Demonstrator Precursor Digital Assistant Interface For Digital Twin Earth (DA4DTE) Project.
\bibliographystyle{IEEEtranDOI}
\bibliography{refs/defs.bib,refs/refs.bib}
%\vspace{-1cm}
\begin{IEEEbiography}[{\includegraphics[width=1in,height=1.25in,clip,keepaspectratio]{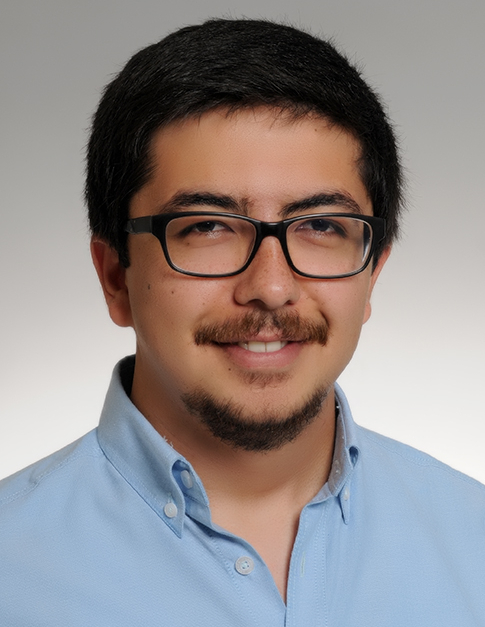}}]{Gencer Sumbul} received his B.S. degree in Computer Engineering from Bilkent University, Ankara, Turkey in 2015 and the M.S. degree in Computer Engineering from Bilkent University in 2018. He is currently a research associate in the Remote Sensing Image Analysis (RSiM) group and pursuing the Ph.D. degree at the Faculty of Electrical Engineering and Computer Science, Technische Universit\"at Berlin, Germany since 2019. His research interests include computer vision, pattern recognition and machine learning, with special interest in deep learning, large-scale image understanding and remote sensing. He is a referee for journals such as the IEEE Transactions on Image Processing, IEEE Transactions on Geoscience and Remote Sensing, the IEEE Access, the IEEE Geoscience and Remote Sensing Letters, the ISPRS Journal of Photogrammetry and Remote Sensing and international conferences such as European Conference on Computer Vision and IEEE International Geoscience and Remote Sensing Symposium.
\end{IEEEbiography} 
% \vfill
%\vspace{-3cm}
\begin{IEEEbiography}[{\includegraphics[width=1in,height=1.25in,clip,keepaspectratio]{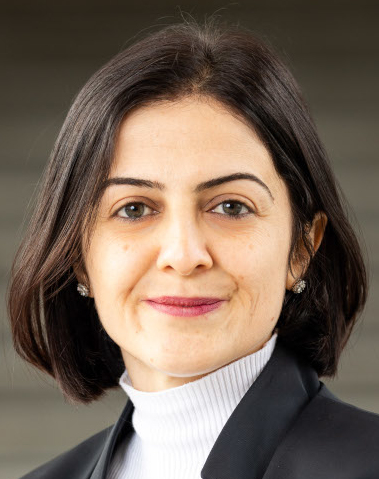}}]{Beg\"{u}m Demir} (S'06-M'11-SM'16) received the B.S., M.Sc., and Ph.D. degrees in electronic and telecommunication engineering from Kocaeli University, Kocaeli, Turkey, in 2005, 2007, and 2010, respectively.

She is currently a Full Professor and the founder head of the Remote Sensing Image Analysis (RSiM) group at the Faculty of Electrical Engineering and Computer Science, TU Berlin and the head of the Big Data Analytics for Earth Observation research group at the Berlin Institute for the Foundations of Learning and Data (BIFOLD). Her research activities lie at the intersection of machine learning, remote sensing and signal processing. Specifically, she performs research in the field of processing and analysis of large-scale Earth observation data acquired by airborne and satellite-borne systems. She was awarded by the prestigious ‘2018 Early Career Award’ by the IEEE Geoscience and Remote Sensing Society for her research contributions in machine learning for information retrieval in remote sensing. In 2018, she received a Starting Grant from the European Research Council (ERC) for her project “BigEarth: Accurate and Scalable Processing of Big Data in Earth Observation”. She is an IEEE Senior Member and Fellow of European Lab for Learning and Intelligent Systems (ELLIS).

Dr. Demir is a Scientific Committee member of several international conferences and workshops, such as: Conference on Content-Based Multimedia Indexing, Conference on Big Data from Space, Living Planet Symposium, International Joint Urban Remote Sensing Event, SPIE International Conference on Signal and Image Processing for Remote Sensing, Machine Learning for Earth Observation Workshop organized within the ECML/PKDD. She is a referee for several journals such as the PROCEEDINGS OF THE IEEE, the IEEE TRANSACTIONS ON GEOSCIENCE AND REMOTE SENSING, the IEEE GEOSCIENCE AND REMOTE SENSING LETTERS, the IEEE TRANSACTIONS ON IMAGE PROCESSING, Pattern Recognition, the IEEE TRANSACTIONS ON CIRCUITS AND SYSTEMS FOR VIDEO TECHNOLOGY, the IEEE JOURNAL OF SELECTED TOPICS IN SIGNAL PROCESSING, the International Journal of Remote Sensing), and several international conferences. Currently she is an Associate Editor for the IEEE GEOSCIENCE AND REMOTE SENSING LETTERS, MDPI Remote Sensing and International Journal of Remote Sensing.
\end{IEEEbiography}
\vfill
\end{document}